%% file: elsarticle-template.tex
\documentclass[review,10pt,doublespace]{elsarticle}

\usepackage{lineno,hyperref}
\usepackage{epsfig}
\usepackage{graphicx}
\usepackage{amsmath,amssymb}
\usepackage[bf]{caption}
\usepackage{cs}
\usepackage{mathsymb}
\usepackage{textcomp}
\usepackage{verbatim}
\usepackage{url}
\usepackage[super]{nth}
\usepackage{colortbl}
\usepackage{subcaption}
\usepackage{multirow}
\usepackage[linesnumbered,algoruled,boxed,lined]{algorithm2e}
\usepackage[noend]{algpseudocode}
\usepackage{wrapfig,lipsum,booktabs}
\usepackage{pbox}
\usepackage{wrapfig}
\graphicspath{{./}{./Fig/}{./Figs/}{./Fig/eps/}{./Fig/pdf/}}

\def\bw{{\boldsymbol w}}

\newcommand{\eg}{\textit{e.g., }}
\newcommand{\ie}{\textit{i.e., }}

\newtheorem{dfn}{Definition}   

\journal{}

\begin{document}

\begin{frontmatter}

\title{Deep Adaptive Feature Embedding with Local Sample Distributions for Person Re-identification }

\author[label1]{Lin Wu \corref{cor1}}
\ead{lin.wu@uq.edu.au}
\author[label2]{Yang Wang}
\ead{wangy@cse.unsw.edu.au}
\author[label3]{Junbin Gao}
\ead{junbin.gao@sydney.edu.au}
\author[label1]{Xue Li}
\ead{xueli@itee.uq.edu.au}

\cortext[cor1]{Corresponding author.}

\address[label1]{ The University of Queensland, Queensland 4072, Australia}
\address[label2]{The University of New South Wales, NSW 2052, Australia}
\address[label3]{Discipline of Business Analytics, The University of Sydney Business School, The University of Sydney, NSW 2006, Australia}


\begin{abstract}
Person re-identification (re-id) aims to match pedestrians observed by disjoint camera views. It attracts increasing attention in computer vision due to its importance to surveillance system. To combat the major challenge of cross-view visual variations, deep embedding approaches are proposed by learning a compact feature space from images such that the Euclidean distances correspond to their cross-view similarity metric. However, the global Euclidean distance cannot faithfully characterize the ideal similarity in a complex visual feature space because features of pedestrian images exhibit unknown distributions due to large variations in poses, illumination and occlusion. Moreover, intra-personal training samples within a local range are robust to guide deep embedding against uncontrolled variations, which however, cannot be captured by a global Euclidean distance. In this paper, we study the problem of person re-id by proposing a novel sampling to mine suitable \textit{positives} (\ie intra-class) within a local range to improve the deep embedding in the context of large intra-class variations. Our method is capable of learning a deep similarity metric adaptive to local sample structure by minimizing each sample's local distances while propagating through the relationship between samples to attain the whole intra-class minimization. To this end, a novel objective function is proposed to jointly optimize similarity metric learning, local positive mining and robust deep embedding. This yields local discriminations by selecting local-ranged positive samples, and the learned features are robust to dramatic intra-class variations. Experiments on benchmarks show state-of-the-art results achieved by our method.
\end{abstract}


\end{frontmatter}


\section{Introduction}\label{sec:intro}

\begin{figure*}[t]
\begin{tabular}{cc}
        \includegraphics[width=2in,height=1.5in]{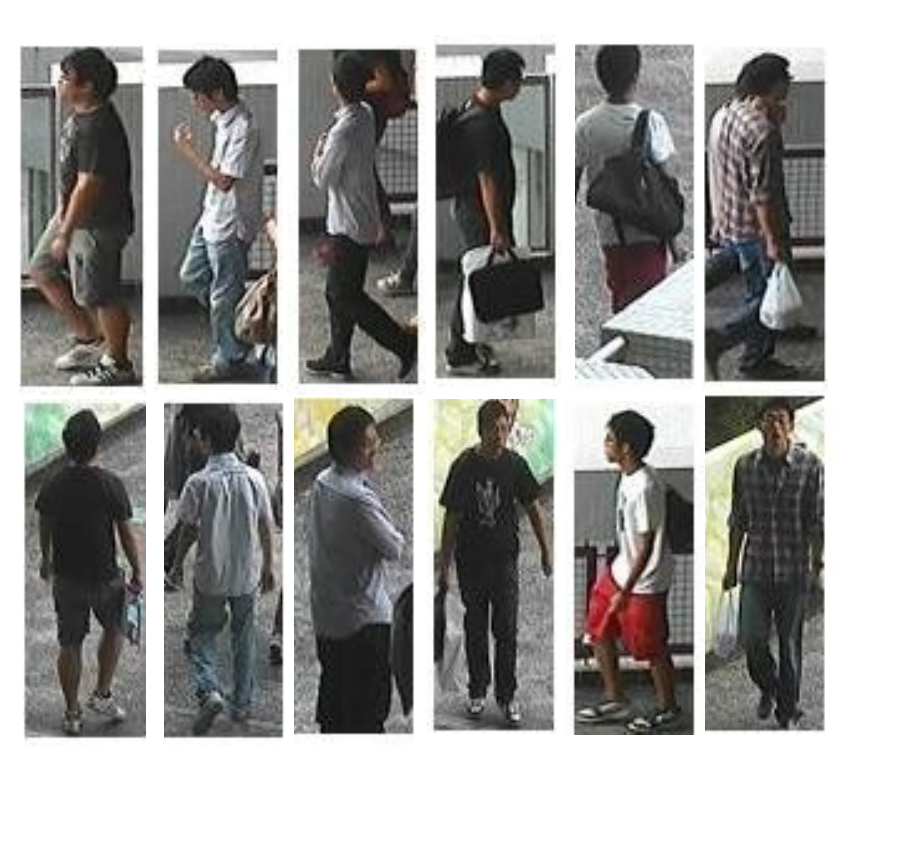}&
        \includegraphics[width=2.5in,height=1.5in]{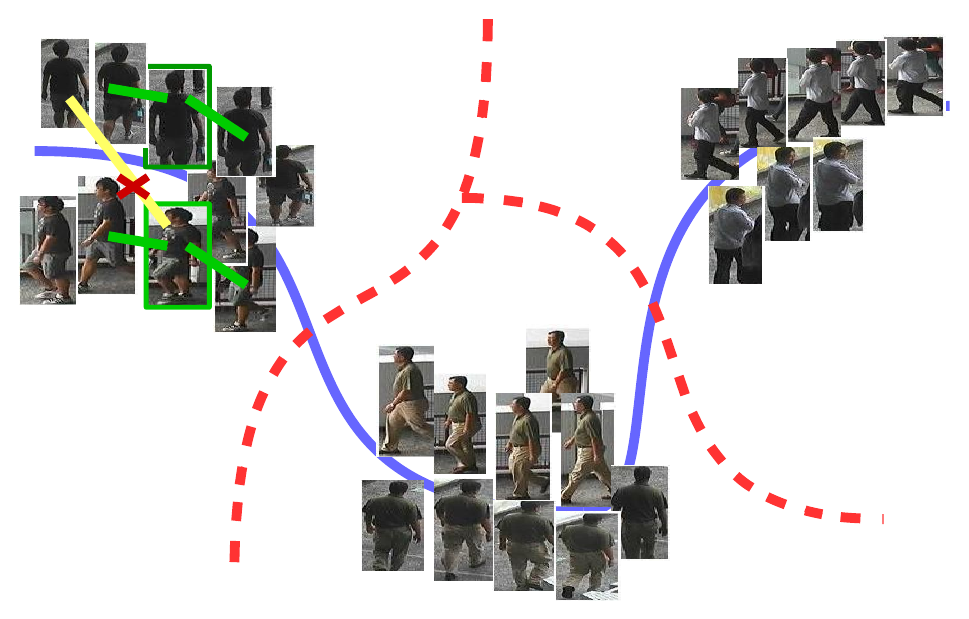}\\
        (a) & (b)
\end{tabular}
\caption{(a) Samples of pedestrian images from the CUHK03 dataset \cite{FPNN}. Each column shows two images of the same identity observed by two disjoint camera views. (b) Highly-curved manifolds of 3 identities. Positive samples in a local range (green lines) should be selected to guide deep embedding while those in large distance (yellow line with cross) should not be sampled to respect the manifold structure.}\label{fig:example}
\end{figure*}

The re-identification (re-id) of individuals across spatially disjoint camera views has attracted tremendous attention in computer vision community due to its practice into security and surveillance systems. Despite years of great efforts, person re-id still remains a challenging task due to its large variations in terms of view points, illuminations and different poses (See examples in Fig.\ref{fig:example} (a)).

Existing approaches to person re-id can be summarized into two categories. The first category focuses on developing robust descriptors to describe a person's appearance against challenging factors (lighting, pose, etc) while preserving identity information \cite{Gheissari2006Person,Farenzena2010Person,Gray2008Viewpoint,MidLevelFilter}. Low-level features such as color \cite{MidLevelFilter,YLXMSijcai16}, texture (Local Binary Patterns \cite{LocallyAlligned,Gheissari2006Person,YXLZpakdd}, and Gabor \cite{Gray2008Viewpoint,YLQ13,YWLXZTNNLS17,YXLQWTIP15,YXLWQMM14,LocallyAlligned}) are commonly used for this purpose. However, direct matching pedestrians based on hand-crafted features is not distinctive and reliable enough to severe changes and misalignment across camera views. The second category \cite{Pedagadi2013Local,LocalMetric,Zheng2011Person,PCCA,Kostinger2012Large,YJLQWPAKDD14,LOMOMetric} comes up with the metric learning problem which is to discriminate distance metrics from training data consisting of cross-camera matched pairs, under which inter-class and intra-class variations of pedestrian samples are maximized and minimized, respectively. They, however, consider feature extraction and metric learning as two independent components, leading to a suboptimal performance. Moreover, such methods focus on optimizing a linear transformation on the input, which has a limited number of parameters and fail to model the higher-order correlations over the original data dimensions.

More recent studies on deep embedding methods \cite{FPNN,DeepReID,JointRe-id,PersonNet,DomainDropout,LocalMetric,DeepRanking,Multi-channel-part} aim at learning a compact feature embedding $f(x)\in \mathbb{R}^d$ from image $x$ via a deep convolutional neural network (CNN). The embedding objective is usually modeled over Euclidean space: the Euclidean distance $D(x_i,x_j)=||f(x_i)-f(x_j)||_2$ between feature vectors should preserve the semantic relationship encoded in pairwise (by contrastive loss \cite{FPNN,DeepReID,JointRe-id,PersonNet}), in triplets \cite{DeepRanking,Multi-channel-part,YXLWQ15,LY17IVC}, or even high-order relationships \cite{StructuredDeepHash,YXLINFSCI,YXLKAIS15,LYJMM13,YLQ13}. Among these methods, hard sample mining is crucial to ensure the quality and the learning efficiency, due to the fact that there are many more easy examples than those meaningful hard examples. Thus, they usually choose hard samples to compute the convenient Euclidean distance in the embedding space. However, these deep embedding methods suffer from inherent limitations: First, they adopt a global Euclidean distance metric to evaluate the hard samples whereas recent manifold learning in person re-id \cite{E-metric} suggests that pedestrian samples are distributed as highly-curved manifolds. Euclidean distance can only be adopted in local range to approximate the geodesic distance via graphical relationship between samples (as illustrated in Fig.\ref{fig:example} (b)). Second, these methods are conditioned on individual samples in term of pairs/triplets to categorize the inputs as depicting either the same or different subjects. Such mapping to a scalar prediction of similarity score based on person identities would make the optimization on CNN parameters over-fitting because the supervision binary similarity labels (0 for dissimilar and 1 for similar) tend to push the scores independently. In practice, the similarity scores of positive and negative pairs live on a 1-D manifold following the distribution on heterogeneous data \cite{PositionDeepMetric}. Finally, when training the CNN with contrastive or triplet loss for embedding, existing methods use the Euclidean distance indiscriminately with all the positive samples. Nonetheless, we observe that selecting positive samples within local ranges (pairs in green lines in Fig.\ref{fig:example} (b)) is critical for training whilst enforce training with the positive samples of long distance may distort the manifold structure (yellow line with red cross in Fig.\ref{fig:example} (b)). Moreover, objective functions defined on triplet loss involve sampling on divergent triplets, which is not necessarily consistent, and thus impedes the convergence rate and training efficiency.

\paragraph{Our Approach} Mitigating the aforementioned issue, in this paper we propose a principled approach to learn a local-adaptive similarity metric, which will be exploited to search for suitable positive samples in a local neighborhood to facilitate a more effective yet efficient deep embedding learning. The key challenge lies in the design of robust feature extraction and the loss function that can jointly consider 1) similarity metric learning; 2) suitable positive sample selection; and 3) deep embedding learning. Existing deep embedding studies \cite{FPNN,PersonNet,DomainDropout,SI-CI,YXLWTIP17,YXLW15,E-metric,GatedCNN} only consider the two later objectives but not jointly with the first important aspect. To this end, we propose a principled approach to train a deep network that transforms the input data into a deep feature space where the local  data distribution structure within classes can be captured. We formulate the feature extractor as a stacked convolutional Restricted Boltzmann Machines (CRBMs) \cite{CRBM} to initialize the parameters that define the mapping from input images to their representation space. We remark that CNN has generic parameterization while in person re-id case, body parts exhibit different visual modalities due to the combinations of view points, poses, and photometric settings \cite{Multiregion,LocallyAlligned,IEEETCYB16}. Thus, a single/generic CNN filter cannot capture the inter-camera variations while some fine-grained information such as ``texture in clothes" and ``bags" are very helpful in reducing intra-personal variations. As such, CRBMs serve as hierarchical feature model to faithfully describe pedestrian samples containing dramatic variations. We formulate the training of CRBMs adaptively to search the suitable positive samples within local range so that it learns locally adaptive metric (instead of global Euclidean distance). Furthermore, to improve training efficiency, we employ variance reduced Stochastic Gradient Descent (SGD) \cite{VRSGD} to share and reuse past stochastic gradients across data samples by exploiting their neighborhood structure.
As shown in Fig.\ref{fig:framework}, the proposed metric yields similarity scores in mini-batch, from which positive samples constituting a hard quadruplet are mined and used to optimize the feature embedding space. The similarity metric learning and embedding learning in the associated CRBMs are jointly optimized via a novel large-margin criterion.

\paragraph{Contributions} The main contributions of our work are four-fold:
\begin{itemize}
\item An improved deep embedding approach is presented to construct a representation amenable to similarity metric computation in person re-identification by jointly optimizing robust feature embedding, local adaptive similarity learning, and suitable positive mining.
\item The proposed method enhances the quality of learned representations and the training efficiency by accessing Euclidean distance of samples in local range w.r.t highly-curved structure. This allows adaptive similarity access in local range and achieves minimization of intra-class variations by local-ranged positive sample mining.
\item We provide alternative to CNN embedding by formulating a stacked CRBMs into local sample structure in deep feature space, and thus enables local adaptive similarity metric learning as well as plausible positive mining.
\item Our method achieves state-of-the-art results on four benchmark datasets: VIPeR \cite{Gray2007Evaluating}, CUHK03 \cite{FPNN}, CUHK01 \cite{GenericMetric}, and Market-1501 \cite{Market1501}.
\end{itemize}

\section{Related Work}\label{sec:related}

\subsection{Metric Learning in Person Re-identification}

Metric learning algorithms have been extensively applied into person re-identification to learning discriminative distance metrics or subspaces for matching persons across views \cite{E-metric,Pedagadi2013Local,SimilaritySpatial,LADF,Xiong2014Person,Zheng2013PAMI,NullSpace-Reid,GenericMetric,Farenzena2010Person,PCCA,Kostinger2012Large,LocalMetric,LOMOMetric}.
They essentially perform a two-stage pipeline where hand-crafted features are extracted for each image, and then a Mahalanobis form metric is learned. This corresponds to a linear projection to map training examples into a discriminative subspace, such that inter-category difference and intra-category similarities are preserved. For instance, Pairwise Constrained Component Analysis (PCCA) \cite{PCCA} is proposed to learn a projection into a low-dimensional space under which the distance between pairs of images respects the desired constraints. Zheng \etal \cite{Zheng2013PAMI} propose the Relative Distance Comparison (RDC) approach to maximize the likelihood of a pair of true matches having a relatively smaller distance than that of a wrongly matched pair in a soft discriminant manner. In \cite{LADF},  Li \etal develop a Locally-Adaptive Decision Function (LADF) that can jointly learn the distance metric and a locally adaptive thresholding rule. Local Fisher Discriminant Analysis (LFDA) is introduced by Pedagadi \etal \cite{Pedagadi2013Local} to learn a subspace to reduce the dimensionality of the extracted high-dimensional features under which the Fisher discriminant criterion is met \cite{LDAFisherVector}. These methods have a common drawback in terms of the separation on feature extraction and metric learning, making their performance limited by the representation power of low-level features. Moreover, they aim to optimize a linear transformation with a limited number of parameters, which cannot model high-order correlations between original data dimensions.

To jointly learn representations and similarity metric for pedestrian samples, deep embedding approaches are developed to allow the interaction between feature extraction and metric learning \cite{FPNN,DeepReID,JointRe-id,PersonNet,DomainDropout,DeepRanking,Multi-channel-part,DeepRDC-Person}. Euclidean distance is the simplest similarity metric, and widely used by current deep embedding methods where Euclidean feature distances directly correspond to the similarity. Similarities can be encoded in pairwise with a contrastive loss \cite{FPNN,JointRe-id,PersonNet}, or a flexible triplet loss \cite{DeepRanking,Multi-channel-part,DeepRDC-Person}. Alternative to Euclidean metric is parametric Mahalanobis metric, representative works \cite{LocalMetric,NullSpace-Reid} minimize the Mahalanobis distance between the positive sample pairs while maximizing the distance between negative pairs. They directly optimize the the Mahalanobis metric for nearest neighbor classification via the strategy of Large Margin Nearest Neighbor \cite{Weinberger2006Distance}. However, Mahalanobis metric learning is also a global linear transformation of the input space that precedes $k$-NN classification using Euclidean distance. Thus, the common drawbacks of Mahalanobis and Euclidean metric is that they are both global and and unable to reflect the heterogeneous feature distribution.

As data samples reside on highly-curved manifold, this indicates the similarity notion should be defined on local Euclidean distance which can be propagated to approximate the whole class structure by enforcing reasonable proximity relationship between samples. In this paper, we present an approach to train a deep neural network to learn deep embeddings from the input space to a discriminative feature space under which the similarity is defined as a function of local range structure within a subject such that the intra-personal variations can be reduced substantially.

\subsection{ Deep Embeddings with Hard Sample Mining in Person Re-identification}

Person re-id is a challenging task in terms of the large intra-personal variations (\eg, viewpoints, poses, occlusion) present in typical surveillance footage. Thus, it requires a fine granularity model to discriminate identities that resemble each other with subtle difference. Deep learning has shown great success in a variety of tasks in image classification \cite{AlexNet,VGG}, human face verification \cite{Face-verify} and frequency domain \cite{CNNpack,Compress-vector}. Inspired by these high-capacity models in deep learning, some deep embedding models have been developed for person re-id to learn representations against visual variations \cite{FPNN,DeepReID,JointRe-id,PersonNet,DomainDropout,LocalMetric,DeepRanking,Multi-channel-part,E-metric,GatedCNN,SI-CI,S-LSTM,DeepRDC-Person}. They commonly learn a feature embedding from images using a deep CNN, and optimize an embedding objective in an Euclidean distance, which should preserve their semantic relationship encoded by pairwise (contrastive loss \cite{SI-CI,FPNN,PersonNet,JointRe-id}), in triplets \cite{DeepRanking,Multi-channel-part,DeepRDC-Person}, or even higher order relationships (\eg a structured loss \cite{StructuredDeepHash}).  An important component of deep embedding approaches is hard sample mining, which is crucial to ensure the learning quality and the efficiency since there would be many easy examples than those meaningful hard examples. In existing deep embedding pipelines of person re-id, hard sample mining is commonly performed to augment a training set progressively with false positive examples with the model learned so far. However, they select false positive/hard negative examples randomly, in which divergent pairs/triplets are not necessarily consistent, and hinders the convergence rate. Moreover, the similarity is defined in a global Euclidean distance, which cannot capture the complex feature structure. To this end, we are motivated to develop a deep neural network to learning a deep transformation from input space to a representation space in which neighborhood structure w.r.t class distributions is adaptively captured. Meanwhile, we characterize the similarity adaptively as function of positive samples in local-ranged feature space, and pursue local large margin as opposed to global.
\section{Deep Feature Embedding with Local Adaptive Similarities}\label{sec:approach}

\begin{figure*}[t]
\centering
\includegraphics[height=3cm]{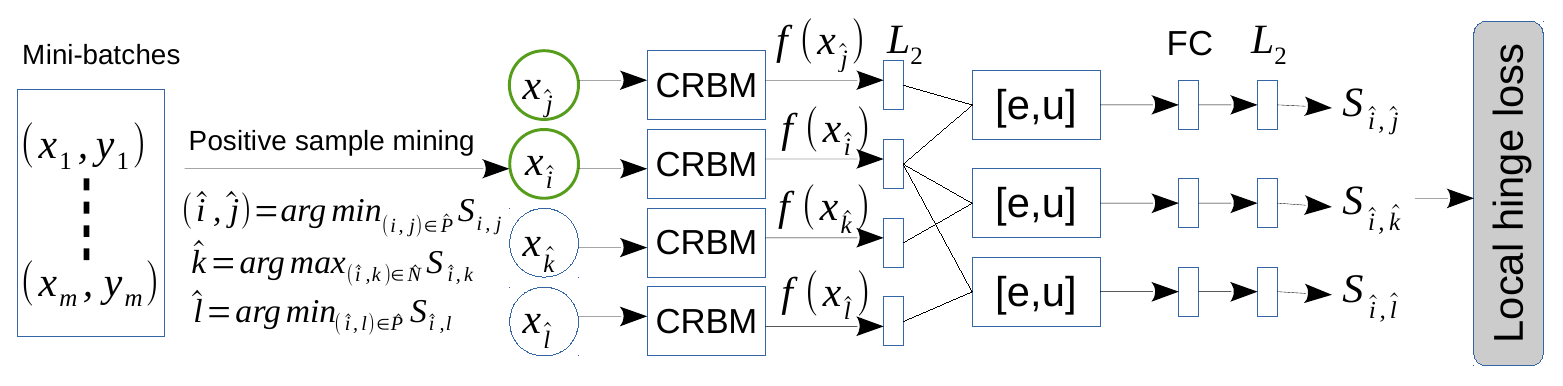}
\caption{The proposed deep embedding framework. For each feature pair ($\boldsymbol f_W(x_i), \boldsymbol f_W(x_j)$) extracted from images $x_i$ and $x_j$ by an embedding function $\boldsymbol f_W(\cdot)$ parameterized by $W$, we first select the most dissimilar positive pair in the mini-batch via $(\hat{i},\hat{j})=\arg\min_{(i,j)\in \bar{P}} S_{i,j}$. where $S_{i,j}$ is the similarity score. Then, a hard quadruplet is constructed by choosing the hard negative $\hat{k}$ and hard positive $\hat{l}$ via $\hat{k}=\arg\max_{(\hat{i},k)\in \bar{N}} S_{\hat{i},k}$ and $\hat{l}=\arg\min_{(\hat{i},l)\in \bar{P}} S_{\hat{i},l}$, respectively. The selected hard quadruplet $[\hat{i},\hat{j},\hat{l},\hat{k}]$ is put through the network to calculate their similarities $S_{\hat{i},\hat{j}}, S_{\hat{i},\hat{k}}, S_{\hat{i},\hat{l}}$ which are optimized under large margin criterion so as to jointly learn similarity metric and adapt their embeddings $[\boldsymbol f(x_{\hat{i}}), \boldsymbol f(x_{\hat{j}}), \boldsymbol f(x_{\hat{k}}), \boldsymbol f(x_{\hat{l}})]$ into local sample distributions. Note that $W$s are parameterized by CRBMs. }
\label{fig:framework}
\end{figure*}

In this section, we present our approach to learn adaptive, deep embeddings that transform the input data with real-values into a representation space such that the intra-class variations can be addressed by selecting suitable positive samples in a local range. Formally, let $X=\{x_i,y_i\}$ be a pedestrian imagery dataset, where $y_i$ is the class/identity label of image $x_i$. Our goal is to jointly learn a deep feature embedding $\boldsymbol f(x)$ from image $x$ into a feature space $\mathbb{R}^d$, and a similarity metric $ S(\boldsymbol f(x_i), \boldsymbol f(x_j)) \in \mathbb{R}^1$, such that the metric can robustly select positive samples adaptive to local range to learn a discriminative feature embedding. Ultimately, the learned features $(\boldsymbol f(x_i), \boldsymbol f(x_j))$ from the set of positive pairs $P=\{(i,j)|y_i=y_j\}$ should be close to each other with a large similarity value $S_{i,j}$, whilst those from the set of negative pairs $N=\{(i,j)|y_i\neq y_j\}$ should be far away with a small similarity score. Importantly, this relationship cannot be reflected in a global Euclidean metric $S_{i,j}$ wherein data samples reside on highly-curved manifolds and Euclidean distance is limited in the local range. To adapt $S_{i,j}$ into local latent structure of feature embeddings, we propose to select suitable positive samples within a local neighborhood to guide deep embedding adaptive to manifold structure.

The overall architecture is shown in Fig.\ref{fig:framework}. Given a mini-batch containing pairs of images represented in the form of $(\boldsymbol f_W(x_i), \boldsymbol f_W(x_j))$, we compute their similarity scores $S_{i,j}$ to select one hard quadruplet from the local set of positive pairs $\bar{P}\in P$, and negative pairs $\bar{N}\in N$ in the batch. Then, each sample in the mined hard quadruplet is fed into four identical Convolutional Restricted Boltzmann Machines (CRBMs) with shared parameters $W$ to extract $d$-dimensional features. To optimize the parameters $W$, a discriminative local loss is applied to similarity scores based on a large margin criterion (Section \ref{ssec:fine-tuning}). Note that we initially use CRBMs \cite{CRBM} to produce features that discovers the structure of complex visual appearance on pedestrian images (Section \ref{ssec:pretraining}). The features from CRBMs can be used to compute similarity scores for the mini-batch samples during a particular forward pass.

\subsection{Joint Similarity Learning and Local Positive Sample Mining }\label{ssec:fine-tuning}

To enforce similarity into learned representations in which intra-class variations among pedestrian samples are minimized by enforcing reasonable proximity relationship, we are motivated to learn a similarity metric to adapt into local structure where each example is designated only a few number of target neighbors of the same class \cite{Weinberger2006Distance,Multi-stage-metric,Dim-reduction2006}. This principle suits to person re-identification datasets where each identity is associated with only a small number of images.

\subsubsection{Adapting to Local Sample Distributions}

Given a feature pair $(\boldsymbol f_W(x_i), \boldsymbol f_W(x_j))$ extracted from images $x_i$ and $x_j$ by an embedding function $\boldsymbol f_W(\cdot)$ parameterized by $W$, we aim to learn a similarity score $y_{i,j}=1$ if $(i,j)\in P$, and $y_{i,j}=0$ if $(i,j)\in N$. Thus, we seek the optimal similarity metric $S^*(\cdot,\cdot)$, and feature embedding parameters $W^*$:
\begin{equation}
[S^*(\cdot,\cdot), W^*]=\arg \min_{S(\cdot,\cdot), W} \frac{1}{|P\cup N|} \sum_{(i,j)\in P\cup N} \mathcal{L}(S(\boldsymbol f_W(x_i), \boldsymbol f_W(x_j)), y_{i,j}),
\end{equation}
where $\mathcal{L}(\cdot)$ is a loss function to be defined later. We will omit the parameter notation $W$ from $\boldsymbol f_W(\cdot)$ in the following for brevity.

The standard Euclidean or Mahalanobis metric is defined based on the feature difference vector $\boldsymbol e=|\boldsymbol f(x_i)- \boldsymbol f(x_j)|$ or its linear transformation. However, these metrics are demonstrated to be suboptimal in a heterogeneous embedding space despite of the CRBM's capability of hierarchical feature extraction. By contrast, similarity metric learning should characterize similarity adaptively into local sample structure in the deep feature space. This knowledge can be utilized to reduce large intra-class visual variations in the resulting representation space. Thus, inspired by \cite{RandomPosition,PositionDeepMetric}, apart from the feature difference vector $\boldsymbol e$, we additionally introduce the feature mean vector $\boldsymbol u=(\boldsymbol f(x_i) + \boldsymbol f(x_j))/2$ to leverage absolute feature position to adapt the metric into local range.

Formally, following the procedure suggested by \cite{PositionDeepMetric}, the features $\boldsymbol f(x_i)$ and $\boldsymbol f(x_j)$ are first normalized onto the unit hypersphere, that is, $||\boldsymbol f(x)||_2=1$, in order to maintain feature compatibility in the computation of their relative and absolute positions encoded by $\boldsymbol e$ and $\boldsymbol u$, respectively. Thereafter, a sequence of nonlinearities of a fully connected layer, an element-wise ReLU function $\max(0,x)$, and a second $\ell_2$-normalization $r(x)=\frac{x}{||x||_2}$ are applied on $\boldsymbol e$ and $\boldsymbol u$, respectively. This process can be formulated as:
\begin{equation}
\begin{split}
&\boldsymbol e =|\boldsymbol f(x_i)- \boldsymbol f(x_j)|, \boldsymbol u=(\boldsymbol f(x_i) + \boldsymbol f(x_j))/2;\\
&\bar{\boldsymbol e}=r(max(0, W_e \boldsymbol e + b_e)), \bar{\boldsymbol u}=r(max(0, W_u \boldsymbol u + b_u));
\end{split}
\end{equation}
where the parameters ($W_e\in \mathbb{R}^{d\times d}$, $b_e \in \mathbb{R}^d$) and ($W_u\in \mathbb{R}^{d\times d}$, $b_u \in \mathbb{R}^d$) are not shared. Then, the vectors $\bar{\boldsymbol e}$ and $\bar{\boldsymbol u}$ are concatenated and fed into a fully connected layer, parameterized by $W_c \in \mathbb{R}^{2d\times d}, b_c \in \mathbb{R}^d$, and the ReLU function, to map to a final similarity score $S_{i,j}=S(\boldsymbol f(x_i), \boldsymbol f(x_j))\in \mathbb{R}^1$, parameterized by $W_s \in \mathbb{R}^{d\times 1}, b_s \in \mathbb{R}^1$. This can be defined as:
\begin{equation}
S_{i,j}=W_s \boldsymbol c + b_s, \boldsymbol c=\max(0,W_c [\bar{\boldsymbol e}; \bar{\boldsymbol u}]^T + b_c).
\end{equation}
Thus, in this way, the seeking of the similarity metric function $S(\cdot,\cdot)$ can be transformed into the joint learning of hyper-parameters $\{W_e, W_u, W_c, W_s, b_e, b_u, b_c, b_s\}$, and $\{W^1,W^2,W^3,b^1,b^2,b^3\}$ for feature embeddings of CRBMs.

\subsubsection{Local Positive Sample Mining}

\begin{figure}[t]
\centering
\includegraphics[height=4cm]{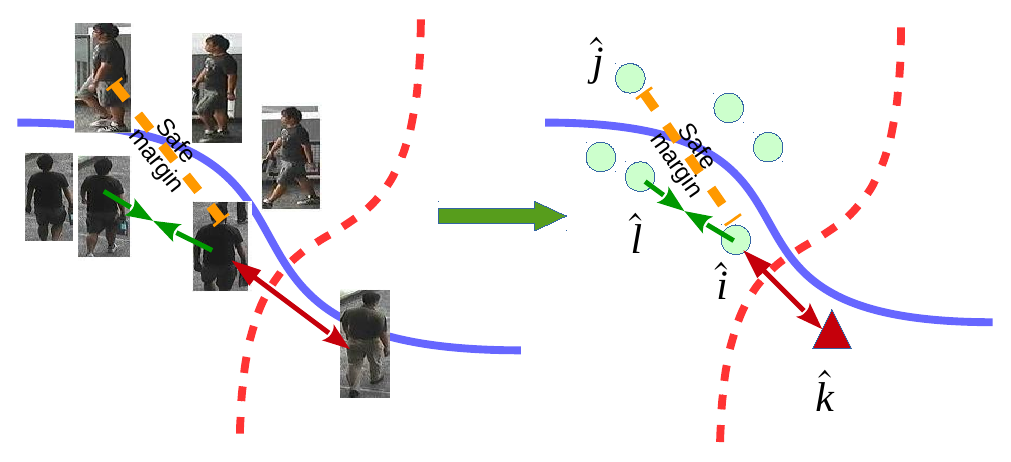}
\caption{Positive sample mining in a local range.}
\label{fig:safe-margin}
\end{figure}

\begin{algorithm}[t]
\KwIn{A mini-batch $X=\{x_i,y_i\}$.}
\KwOut{One hard quadruplet samples.}
Input the images into the CRBMs to obtain their features $\boldsymbol f(x_i)$, and compute their similarity scores $S_{i,j}=S(\boldsymbol f(x_i),\boldsymbol f(x_j))$\;
Mine the most dissimilar positive pair in the mini-batch: $(\hat{i},\hat{j})=\arg\min_{(i,j)\in \bar{P}} S_{i,j}$\;
Choose the hard negative satisfying $\hat{k}=\arg\max_{(\hat{i},k)\in \bar{N}} S_{\hat{i},k}$\;
Choose the hard positive satisfying $\hat{l}=\arg\min_{(\hat{i},l)\in \bar{P}} (S_{\hat{i},l} > S_{\hat{i}, k})$\;
\Return{One hard quadruplet $[\hat{i},\hat{j},\hat{l},\hat{k}]$ }\;
\caption{Local positive sample mining.}
\label{alg:positive_mining}
\end{algorithm}

To optimize all these parameters, we need to choose an appropriate loss function $\mathcal{L}(\cdot)$, in which the pre-trained multi-layer network initialized by CRBMs are optimized to learn deep embeddings with adaptive local sample structure. One possible solution is to cast the problem as a binary classification problem as some deep embedding methods \cite{FPNN,PersonNet,JointRe-id}. However, the binary similarity labels $y_{i,j}\in \{0,1\}$ tend to independently push the scores towards two single points. While some study \cite{PositionDeepMetric} has shown that the similarity scores of positive and negative pairs live on a 1-D manifold following some irregular distribution on data space. This motivates us to design a loss function that is able to separate the similarity distribution in a local range along its manifold. One option is to impose the Fisher criterion \cite{LDA,DeepFisherNetworks,LDAFisherVector,Pedagadi2013Local} on the similarity scores, which is to maximize the ratio between the interclass and intraclass scatters of scores. However, the optimality of Fisher criteria relies on the assumption that the data of each is of Gaussian distribution, which is not satisfied in our case.

To this end, we propose a loss function that approximately maximizes the margin between the positive and negative similarity distribution in a local range to reduce the intra-class variations. Specifically, we select a hard quadruplet from each random mini-batch during each forward pass. To constitute the hard quadruplet, we first select the most dissimilar positive pair in the batch via $(\hat{i},\hat{j})=\arg\min_{(i,j)\in \bar{P}} S_{i,j}$, which indicates their similarity score is most likely to cross the ``\textbf{safe margin}'', and move towards the negative similarity distribution in the local range (see the illustration in Fig.\ref{fig:safe-margin}). Then, we construct the hard quadruplet by choosing the hard negative $\hat{k}$ and hard positive $\hat{l}$ w.r.t each anchor sample $\hat{i}$, via $\hat{k}=\arg\max_{(\hat{i},k)\in \bar{N}} S_{\hat{i},k}$ and $\hat{l}=\arg\min_{(\hat{i},l)\in \bar{P}} (S_{\hat{i},l} > S_{\hat{i}, k})$, respectively. Mining the hard negative $\hat{k}$ w.r.t the anchor image $\hat{i}$ is to ensure the correct relative distances between positive and negative pairs, and thus the sample $\hat{k}$ is push away from the safe margin. On the other hand, we choose the positive samples that have larger similarity scores than the hardest negative, and then mine the hardest one amongst these chosen positives as adaptive local positive samples. This is to preserve the local manifold structure by pushing the positive sample towards the safe margin.

With this hard quadruplet $[\hat{i},\hat{j},\hat{l},\hat{k}]$, we can define the suitable positive samples adaptively within each subject, meanwhile their hard negatives are also involved in case the positive ones are too easy or too hard to be mined. Finally, we design the objective function by discriminating the local similarity distributions under the large margin criterion:
\begin{equation}\label{eq:obj}
\begin{split}
&\min \mathcal{L}=\sum_{\hat{i},\hat{j}} \left(\varepsilon_{\hat{i},\hat{j}}+ \eta_{\hat{i},\hat{j}} \right), \\
& s.t.: \forall (\hat{i},\hat{j}), \max \left(0, \alpha_1+ S_{\hat{i},\hat{k}} - S_{\hat{i},\hat{j}} \right)\leq \varepsilon_{\hat{i},\hat{j}}, \max \left(0, \alpha_2 + S_{\hat{i},\hat{k}} - S_{\hat{i},\hat{l}} \right)\leq \eta_{\hat{i},\hat{j}}, \\
& (\hat{i},\hat{j})=\arg\min_{(i,j)\in \bar{P}} S_{i,j}, \hat{k}=\arg\max_{(\hat{i},k)\in \bar{N}} S_{\hat{i},k}, \hat{l}=\arg\min_{(\hat{i},l)\in \bar{P}} S_{\hat{i},l}, \varepsilon_{\hat{i},\hat{j}} \geq 0, \eta_{\hat{i},\hat{j}} \geq 0,
\end{split}
\end{equation}
where $\varepsilon_{\hat{i},\hat{j}}$, $\eta_{\hat{i},\hat{j}}$ are the slack variables, $\alpha_1$ and $\alpha_2$ are the enforced margins. We adopt the different margin thresholds to determine the balance of two terms in our loss function. Specifically, we require that the margin between the pairs w.r.t the same probe ($S_{\hat{i},\hat{k}} - S_{\hat{i},\hat{j}}$) should be large enough to enlarge the inter-class variations. And the term of local positive mining as opposed to the safe margin could hold smaller margin to preserve a relatively weak constraint on local manifold structure. Thus, $\alpha_1$ is set to be larger than $\alpha_2$. In this sense, the proposed quadruplet loss not only maintains the correct relative distance between positive and negative pairs but also preserves the local manifold structure by mining positive samples in a local range. The procedure of mining the hard quadruplet is illustrated in Fig.\ref{fig:safe-margin} and summarized in Algorithm \ref{alg:positive_mining}.

\subsection{Feature Extraction using Convolutional RBMs}\label{ssec:pretraining}

In this paper, we employ CRBM \cite{CRBM} to extract features from each sample. The weights in CRBM between hidden layers and visible layers (corresponding to input data, such as image pixels) are shared among all locations in an image. Capturing 2-D structure of images in this way allows weights that detect a given feature to be replicated across locations, and thus redundancy can be reduced to make it scalable to realistic full-sized images.

\begin{figure}[t]
\centering
\includegraphics[height=3.5cm]{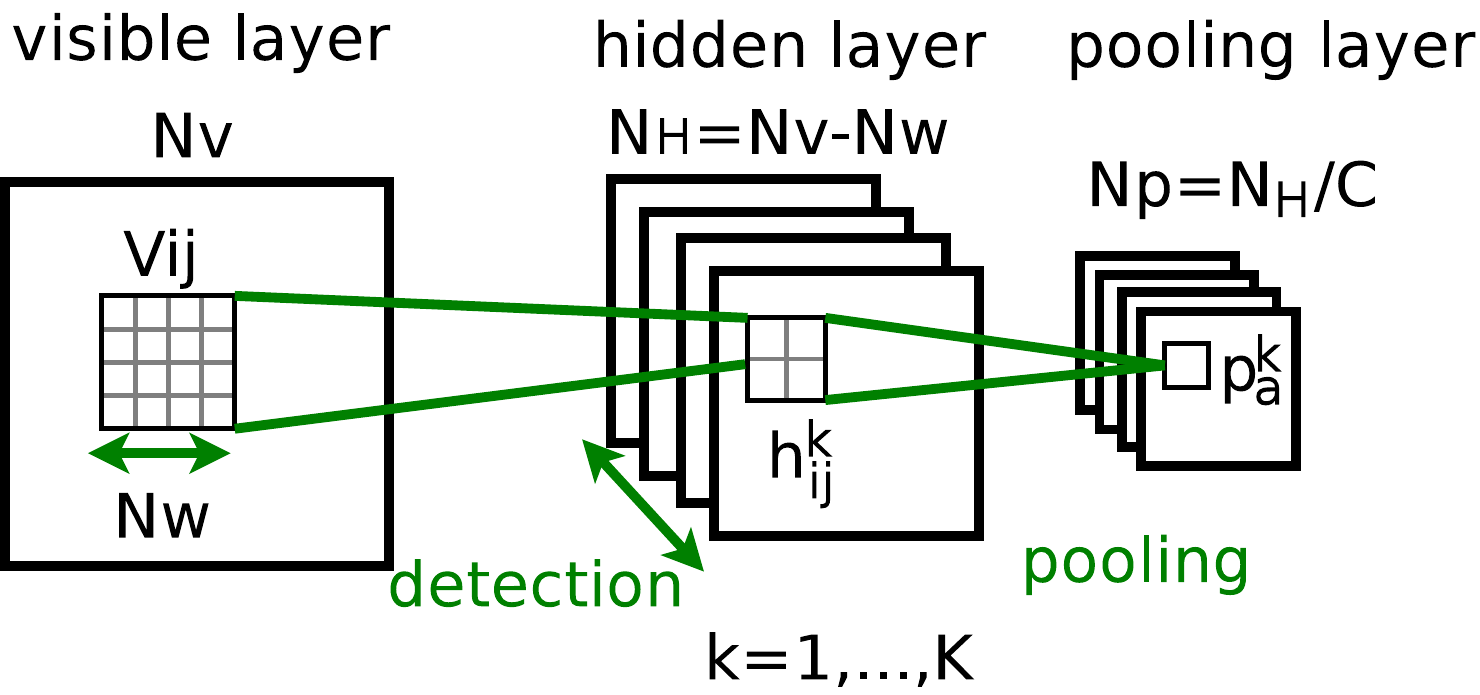}
\caption{A block of convolutional RBM with probabilistic max-pooling \cite{CRBM}. See text for details.}
\label{fig:CRBM}
\end{figure}

The basic CRBM consists of two layers: an input layer $V$ and a hidden layer $H$ with real-valued visible input nodes $\mathbf{v}$ and binary-valued hidden nodes $\mathbf{v}$. The visible input nodes can be viewed as intensity values in the image of $N_V\times N_V$ pixels, and the hidden nodes are manipulated in 2-D configurations, that it, \ie $\mathbf{v}\in \mathcal{R}^{N_V \times N_V}$ and $\mathbf{h}\in \{0,1\}^{N_H\times N_H}$. As illustrated in Fig. \ref{fig:CRBM}, a CRBM block consists of three sets of parameters: (1) $K$ convolution filter weights between a hidden node and a subset of visible nodes where each filter perceives $N_W \times N_W$ pixels (\ie $W^k \in \mathcal{R}^{N_W\times N_W}$, $k=1,\ldots,K$); (2) hidden biases $b^k \in \mathcal{R}$ that are shared among hidden nodes; (3) visible bias $c\in \mathcal{R}$ that is shared among visible nodes. After convolutional detection (filtering), a technique of probabilistic max-pooling \cite{CDBN,CRBM,CDBN-face} is used to incorporate local translation invariance. Probabilistic max-pooling in CRBM acts as max-pooling like behavior, while enabling probabilistic inference (such as bottom-up and top-down inference) scalable to full-size images. Specifically, the probabilistic max-pooling CRBM with real-valued visible inputs can be defined as:

\begin{equation}\label{eq:energy_function}\small
\begin{split}
&P(\mathbf{v},\mathbf{h})=\frac{1}{Z} \exp(-E(\mathbf{v},\mathbf{h}));\\
&E(\mathbf{v},\mathbf{h})=-\sum_{k=1}^K \sum_{i,j=1}^{N_H}\sum_{r,s=1}^{N_W} h_{ij}^k W_{rs}^k v_{i+r-1,j+s-1} + \sum_{i,j=1}^{N_V}\frac{1}{2} v_{ij}^2\\
& - \sum_{k=1}^K b_k \sum_{i,j=1}^{N_H}h_{ij}^k- c\sum_{i,j=1}^{N_V} v_{ij}; s.t. \sum_{(i,j)\in B_{\alpha}} h_{ij}^j \leq 1, \forall k,\alpha.
\end{split}
\end{equation}
where $B_{\alpha}$ refers to a $C\times C$ block of locally neighboring (\eg $2\times2$) hidden units $h_{ij}^k$ that are pooled to a pooling node $p_{\alpha}^k$. Then we discuss sampling the detection layer $H$ and the pooling layer $P$ given the visible layer $V$. Filter $k$ receives the following bottom-up signals from layer $V$: $I(h_{ij}^k)\overset{\Delta}{=} b_k + (\tilde{W}^k \ast v)_{ij}$, where $\ast$ denotes convolution, $\tilde{W}$ denotes flipping the original filter $W$ in both upside-down and left-right directions. With the energy function in \eqref{eq:energy_function}, suppose $h_{ij}^k$ is a hidden unit contained in block $\alpha$ (\ie $(i,j)\in B_{\alpha}$) the conditional probabilities can be computed as follows:

\begin{equation}\label{eq:con_p}\small
P(v_{ij}=1|\mathbf{h})=\mathcal{N}\left(\left(\sum_{k}W^k  \ast  h^k\right)_{ij}+c, 1\right);P(h_{ij}^k=1|\mathbf{v})=\frac{\exp(I(h_{ij}^k))}{1+\sum_{(i',j')\in B_{\alpha}} \exp(I(h_{i'j'}^k))}
\end{equation}
where $\mathcal{N}(\cdot)$ is a normal distribution. The pooling node $p_{\alpha}^k$ is a stochastic random variable that is defined as $p_{\alpha}^k \overset{\Delta}{=} \sum_{(i,j)\in B_{\alpha}} h_{ij}^k$, and thus the marginal posterior can be written as a soft-max function:

\begin{equation}\label{eq:CRBM-obj}
P(p_{\alpha}^k=1| \mathbf{v})=\frac{\sum_{(i',j')\in B_{\alpha}} \exp(I(h_{i'j'}^k))}{1+\sum_{(i',j')\in B_{\alpha}} \exp(I(h_{i'j'}^k))}.
\end{equation}

\begin{figure}[t]
\centering
\includegraphics[height=3cm]{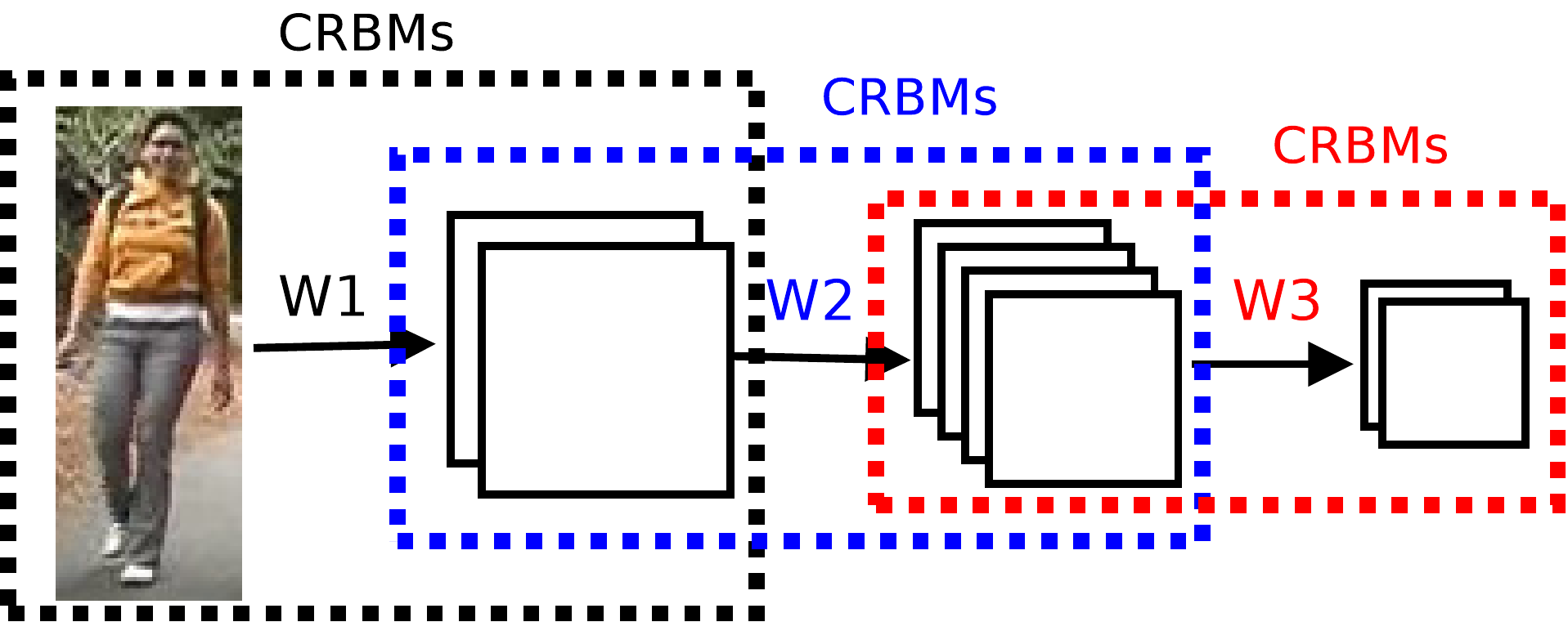}
\caption{Pre-training using three stacked convolutional RBMs \cite{CDBN} in which feature activations of one CRBM are treated as data by the next CRBM. See text for details.}
\label{fig:pretraining}
\end{figure}

We train the RBM parameters by using the contrastive divergence optimization which allows to estimate an approximate gradient efficiently \cite{Hinton-CD}. Since the model is highly over-complete, an appropriate regularization is required to prevent the model from learning trivial feature representations. To this end, a sparsity regularization \cite{CDBN,CRBM} is added into the log-likelihood objective to encourage each hidden unit group to have a mean activation close to a small constant. To increase its expressive power, CRBMs are stacked to form convolutional deep belief network (CDBN) \cite{CDBN}, a hierarchical generative model for full-sized images. CDBN  consists of several max-pooling CRBMs stacked on top of one another, and the network defines an energy function by summing together the energy functions for all of the individual pairs of layers (\ie $E=\sum_{l=1}E(\mathbf{h}^l,\mathbf{h}^{l+1})$ where $\mathbf{h}^0 =\mathbf{v}$). Training CDBN is accomplished with greedy, layer-wise procedure, that is, once a given layer is trained, its weights are frozen, and its activations are used as inputs to the next layer. Once training is completed, hierarchical representations for each image $\boldsymbol f_W(x)$ ($b$ is omitted) can be generated with trainable parameters $W=\{W^1,W^2,W^3\}$, $b=\{b^1,b^2,b^3\}$, namely feature embedding function $\boldsymbol f_W(\cdot)$. Please refer to Appendix for details.

\section{Improvement on Training Efficiency with Variance Reduced SGD}

To improve training efficiency, we adopt variance reduced SGD \cite{VRSGD} to reuse past gradients across data samples with respect to their neighborhood structure. Generally, given a convex loss $\mathcal{L}$, and a $\mu$-strongly convex regularizer $\Omega$, the training of a network aims at finding a parameter vector $\bw$ which minimizes the empirical expectation:

\begin{equation}\label{eq:opt}
\bw^*=\arg\min_\bw f(\bw), f(\bw)=\frac{1}{\mathtt{N}}\sum_{n}^\mathtt{N} f_n(\bw);
f_n(\bw):=\mathcal{L}(\bw,(x_n,y_n)) +\Omega(\bw).
\end{equation}

Steepest descent can find the minimizer $\bw^*$, whereas requires repeated computations of full gradients $f'(\bw)$, which is prohibitive for massive datasets. Stochastic Gradient Descent (SGD) is a widely used alternative in the context of large-scale learning \cite{SGD2010}, which updates only involving $f_n'(\bw)$ for an index $n$ chosen uniformly at random, offering an unbiased gradient estimate. However, studies have shown that SGD has property of slow convergence due to its rate of $O(1/t)$ ($t$ is number of epochs), making training inefficient. Recent findings \cite{DualSGD} show that finite sum of $f(\cdot)$ allows for significantly faster convergence, and it is possible to obtain linear convergence with geometric rates by introducing corrections that ensure convergence for constant learning rates.

\subsection{Memorization algorithms}
Recall the optimization problem in \eqref{eq:opt}, a family of SGD algorithms commonly generates an iterate sequence $\bw^t$ with updates as:
\begin{equation}\label{eq:update}\small
\bw^+ = \bw-\gamma g_n (\bw), g_n(\bw)= f'_n(\bw)- \varepsilon_n, \bar{\varepsilon}_n:=\varepsilon_n-\bar{\varepsilon},
\end{equation}
where $\bar{\varepsilon}:=\frac{1}{\mathtt{N}}\sum_{n}^\mathtt{N} \varepsilon_n$. $\bw$ is the current and $\bw^+$ is the new parameter vector, and $\gamma$ is the step size. $\bar{\varepsilon}_n$ are variance correction terms such at $\bE [\bar{\varepsilon}_n]$=0, which guarantees unbiasedness $\bE [g_n(\bw)]=f'(\bw)$. This is to define updates of asymptotically vanishing variance, \ie $g_n(\bw)\rightarrow 0$ as $\bw \rightarrow \bw^*$, which requires $\bar{\varepsilon}_n \rightarrow f'(\bw^*)$. Herein the \emph{memory} $\varepsilon_n$ can be updated by different algorithms \cite{SAGA}. For instance, the SAGA algorithm \cite{SAGA} maintains variance corrections $\varepsilon_n$ by using the update rule $\varepsilon_n^+ = f'_n(\bw)$ for selected index $n$, and $\varepsilon_j^+=\varepsilon_n$ for $j\neq n$. Thus, SAGA reuses the stochastic gradient $f'(\bw)$ computed at step $t$ to update $\bw$ and $\bar{\varepsilon}$. A variant of SAGA, denoted as $q$-SAGA, updates $q\geq 1$ randomly chosen $\varepsilon_n$ variables at each iteration. And the corrections can be controlled to be $\mathtt{N}/q$ at the expense of additional gradient computations. SVRG \cite{AccelerateSGD} is a randomization framework by fixing $q>0$ and drawing in each iteration $r \sim $ Uniform [0;1). If $r<q/\mathtt{N}$, a complete update $\varepsilon_n^+=f'(\bw) (\forall j)$ is performed, otherwise they are left unchanged.

\subsection{Sharing Gradient Memory with Neighborhoods}

In this paper, we employ a neighborhood based gradient memorization method, N-SAGA \cite{VRSGD} which is motivated by SAGA and SVRG to update $\varepsilon_n$ from selected neighborhood data points with respect to $n$.
\begin{dfn}
The q-memorization algorithms evolve iterates $\bw$ according to \eqref{eq:update} and select in each iteration a random index set $I$ of memory locations to update via:
\begin{equation}\label{eq:N-SAGA}
\varepsilon_j^+ := \begin{cases}
f'_j(\textbf{w}), if ~~j\in I\\
\varepsilon_j, otherwise
\end{cases}
\end{equation}
such that any $j$ has the same probability of $q/\mathtt{N}$ of being updated, \ie $\forall j, \sum_{j\in I} \bP\{I\}=q/\mathtt{N}$.
\end{dfn}

N-SAGA makes use of a neighborhood system $\mathbb{N}_n=\{1,\ldots,\mathtt{N}\}$ and selects neighborhoods uniformly, \ie $\bP\{\mathbb{N}_n\}=\frac{1}{\mathtt{N}}$. Using neighborhoods for sharing gradients between close-by data points can avoid an increase in gradient computations but at the expense of an approximation bias. To this end, we employ two types of quantities. First, the gradient memory $\varepsilon_n$ is defined by using \eqref{eq:N-SAGA}, and the shared gradient memory state $\beta_n$ is used in a modified update rule in \eqref{eq:update}: $\bw^+= \bw-\gamma (f'_n(\bw) -\beta_n +\bar{\beta})$. Assume an index $j$ is selected for the weight update, then we generalize \eqref{eq:N-SAGA} as
\begin{equation}\label{eq:G-N-SAGA}\small
\beta_j^+ := \begin{cases}
f'_n(\textbf{w}), if ~~j\in \mathbb{N}_n\\
\beta_j, otherwise
\end{cases};
\bar{\beta}:=\frac{1}{\mathtt{N}}\sum_{n}^\mathtt{N} \beta_n, \bar{\beta}_n:= \beta_n-\bar{\beta}.
\end{equation}
In \cite{VRSGD}, a proof is given to show that the error can be controlled in a small value: $||\varepsilon_n-\beta_n||^2<\epsilon_n$. Euclidean distances can be used as the metric for defining neighborhoods while standard approximation methods for finding nearest neighbors can also be used.

\section{Relation to Neighborhood Models}

\subsection{Relation to Triplet Relationship based Embedding}

Existing deep metric learning approaches to person re-identification stem from contrastive loss \cite{FPNN,DeepReID,JointRe-id,PersonNet,DomainDropout} and triplet loss \cite{LocalMetric,DeepRanking,Multi-channel-part}. They have the same outline, and for simplicity we use triplet loss as illustration. In a typical triplet training framework, triplet images consisting of a seed example, a positive and negative example to the seed are fed into three network models with shared parameter set to compute their representations $\boldsymbol f (x_i)$, $\boldsymbol f (x_j)$, and $\boldsymbol f (x_k)$ ($i=1,\ldots,M$). Triplet loss demands the distance between mismatched pairs and matched pairs be larger than a pre-defined margin $\alpha\in \mathbb{R}$:
\begin{equation}\label{eq:triplet_loss}\small
\mathcal{L}_{triplet} (W)= \frac{1}{M} \sum_{i=1}^M  \left\{||\boldsymbol f(x_i)-\boldsymbol f(x_k)||_2^2 - ||\boldsymbol f(x_i)- \boldsymbol f(x_j)||_2^2 + \alpha \right\}_+,
\end{equation}
where $\{\cdot\}_+$ denote the hinge function, and $W$ the parameter set of the deep embeddings into the representation space. However, training the deep network with contrastive or triplet loss for embedding uses the global Euclidean distance, which cannot faithfully characterize the true feature similarity in a complex visual feature space. Moreover, penalizing individual pairs or triplets does not take into account of local neighborhood structure, and different combinations of triplets are shown to be not necessarily consistent, hindering the convergence rate. In contrast, modeling local similarity distributions by mining positive samples within classes in the representation space can ensure its adaptation to highly-curved manifolds, and thus effectively reduce intra-class variations.

Our local distribution loss (Eq \eqref{eq:obj}) is an augmentation form of triplet loss by providing more triplet relationship. For each hard quadruplet $[\hat{i},\hat{j},\hat{l},\hat{k}]$ we have selected, it can be reformulated to have two pairs of triplets:
\begin{equation}\small
\begin{split}
&\widehat{\mathcal{L}} = \sum \left\{|| \boldsymbol f(x_{\hat{i}}) - \boldsymbol f(x_{\hat{k}}) ||_2^2 - || \boldsymbol f(x_{\hat{i}}) - \boldsymbol f(x_{\hat{l}}) ||_2^2 + \alpha + ||  \boldsymbol f(x_{\hat{i}}) - \boldsymbol f(x_{\hat{k}}) ||_2^2 - ||   \boldsymbol f(x_{\hat{i}}) - \boldsymbol f(x_{\hat{j}}) ||_2^2 + \alpha\right\}_+\\
& = \left\{ 2|| \boldsymbol f(x_{\hat{i}}) - \boldsymbol f(x_{\hat{k}}) ||_2^2 -  || \boldsymbol f(x_{\hat{i}}) - \boldsymbol f(x_{\hat{l}}) ||_2^2 - || \boldsymbol f(x_{\hat{i}}) - \boldsymbol f(x_{\hat{j}}) ||_2^2 + 2\alpha \right\}_+.
\end{split}
\end{equation}

\subsection{Relation to Neighborhood Component Analysis}
Neighborhood component analysis (NCA) and its extensions \cite{NCA,NonlinearEmbed} are designed to have an objective to maximize the expected number of correctly classified data samples on the labeled training data. They essentially learn a linear/non-linear transformation that transforms input data into a lower-dimensional space to make the $K$-nearest neighbor perform well. The NCA objective is defined as

\begin{equation}\label{eq:NCA}
\mathcal{L}_{NCA}=\frac{1}{\mathtt{N} }\sum_{n=1}^\mathtt{N} -\log \frac{\sum_{n': C(\boldsymbol f(x_{n'}))=C(\boldsymbol f(x_n))} e^{-|| \boldsymbol f(x_n) - \boldsymbol f(x_{n'})||_2^2}}{\sum_{n'=1}^\mathtt{N}  e^{-||\boldsymbol f(x_n) - \boldsymbol f(x_{n'})||_2^2}},
\end{equation}
where $C(x)$ is the class membership of $x$. NCA attempts to model the distribution of data samples by preserving its $K$-nearest neighbors with respect to each data sample. However, this formulation does not address the concern on local range feature structure. Even though we maintain a neighborhood structure, for each example, a naive retrieval to obtain nearest neighbors would lead to completely different classes with high probability.

\section{Experiments}\label{sec:exp}

\begin{figure}[t]
\centering
\includegraphics[height=3.3cm]{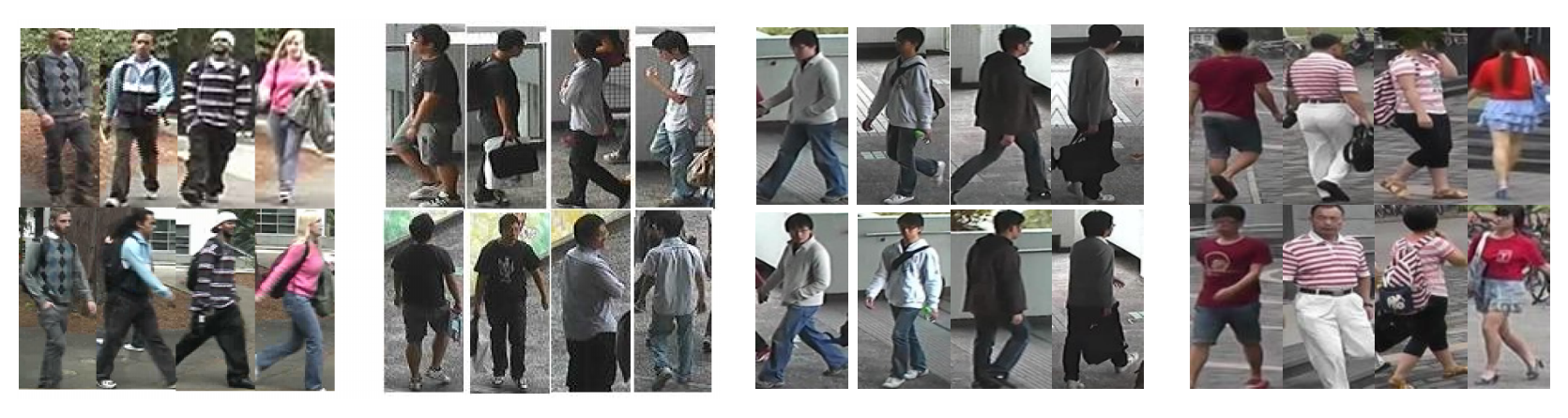}
\caption{Examples from person re-identification datasets. From left to right: VIPeR, CUHK03, CUHK01, and Market-1501.}
\label{fig:dataset}
\end{figure}

\subsection{Data sets}

We perform experiments on four benchmarks: VIPeR  \cite{Gray2007Evaluating}, CUHK03 \cite{FPNN}, CUHK01 \cite{GenericMetric}, and Market-1501 \cite{Market1501}. Examples from these datasets are shown in Fig. \ref{fig:dataset}.

\begin{itemize}
\item The \textbf{VIPeR} data set \cite{Gray2007Evaluating} contains $632$ individuals taken from two cameras with arbitrary viewpoints and varying illumination conditions. The 632 person's images are randomly divided into two equal halves, one for training and the other for testing.

\item The \textbf{CUHK03} data set \cite{FPNN} includes 13,164 images of 1360 pedestrians. The whole dataset is captured with six surveillance camera. Each identity is observed by two disjoint camera views, yielding an average 4.8 images in each view. This dataset provides both manually labeled pedestrian bounding boxes and bounding boxes automatically obtained by running a pedestrian detector \cite{DetectionPAMI}. In our experiment, we report results on labeled data set. The dataset is randomly partitioned into training, validation, and test with 1160, 100,
and 100 identities, respectively.

\item The \textbf{CUHK01} data set \cite{GenericMetric}  has 971 identities with 2 images per person in each view. We report results on the setting where 100 identities are used for testing, and the remaining 871 identities used for training, in accordance with FPNN \cite{FPNN}.

\item The \textbf{Market-1501} data set \cite{Market1501} contains 32,643 fully annotated boxes of 1501 pedestrians, making it the largest person re-id dataset to date. Each identity is captured by at most six cameras and boxes of person are obtained by running a state-of-the-art detector, the Deformable Part Model (DPM) \cite{MarketDetector}.  The dataset is randomly divided into training and testing sets, containing 750 and 751  identities, respectively.

\end{itemize}

\subsection{Experimental Settings}\label{ssec:setting}

Each image in training set is resized to be $150\times 150$ for the visible layer. We process each image by whitening their pixel intensity values, and extracting LBP and Gabor as inputs. LBP can generate 59 dimensional binary vector at each pixel location. Gabor filters are applied on each pixel to extract features invariant to viewpoint and pose \cite{Gray2008Viewpoint}. For each layer of CDBN, we need to set the size of filters, number of filters, and max-pooling region size. The first layer consists of 40 groups ($K=40$) of $12\times 12$ pixel filters ($N_W\times N_W = 12 \times 12$), while the second and third layer consists of 100 groups of $10\times 10$ filters and 40 groups of $6\times 6$ filters, respectively. The deep network of CDBN is optimized on a single GPU GTX 980 using the code provided by \cite{BinaryLatent} \footnote{https://github.com/gwtaylor/imCRBM}. In training CRBMs, each layer was greedily pretrained 50 epoches through the entire training set. The weights were updated using a learning rate of 0.1, momentum of 0.9, and a weight decay of $0.002\times$ weight $\times$ learning rate. The weights were initialized with small random values sampled from a zero-mean normal distribution with variance 0.01. The margins are set to be $\alpha_1=1$, $\alpha_2=0.5$. Activations of three layers (after pooling) are concatenated to be the feature vector.

For the evaluation protocol, we adopt the widely used single-shot modality to allow extensive comparison. Each probe image is matched against the gallery set, and the rank of the true match is obtained. The rank-$r$ recognition rate is the expectation of the matches at rank $r$, and the cumulative values of the recognition rate at all ranks are recorded as the one-trial Cumulative Matching Characteristic (CMC) results. This evaluation is performed ten times, and the average CMC results are reported.

\subsection{Architecture Analysis}

\subsubsection{Local Positive Sample Mining}
\begin{figure*}[t]
   \begin{tabular}{ccc}
  \hspace{-1cm}   \includegraphics[height=3.6cm]{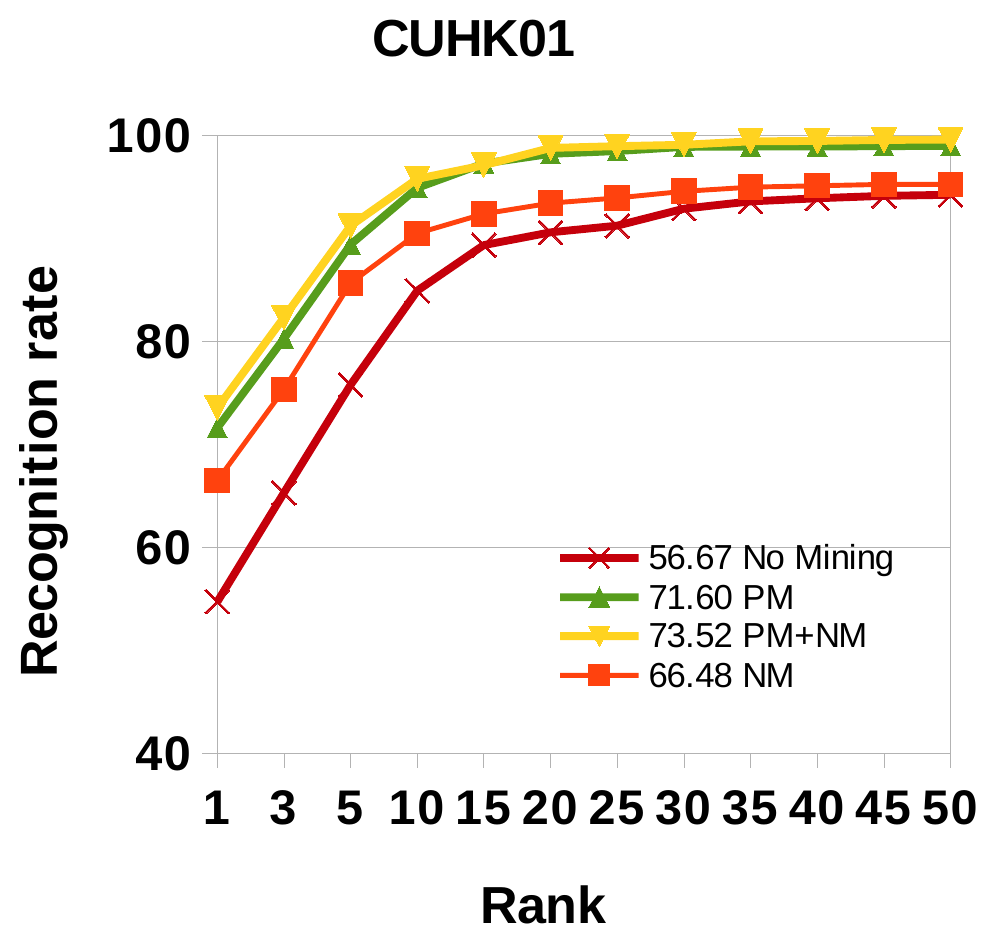}
	\includegraphics[height=3.6cm]{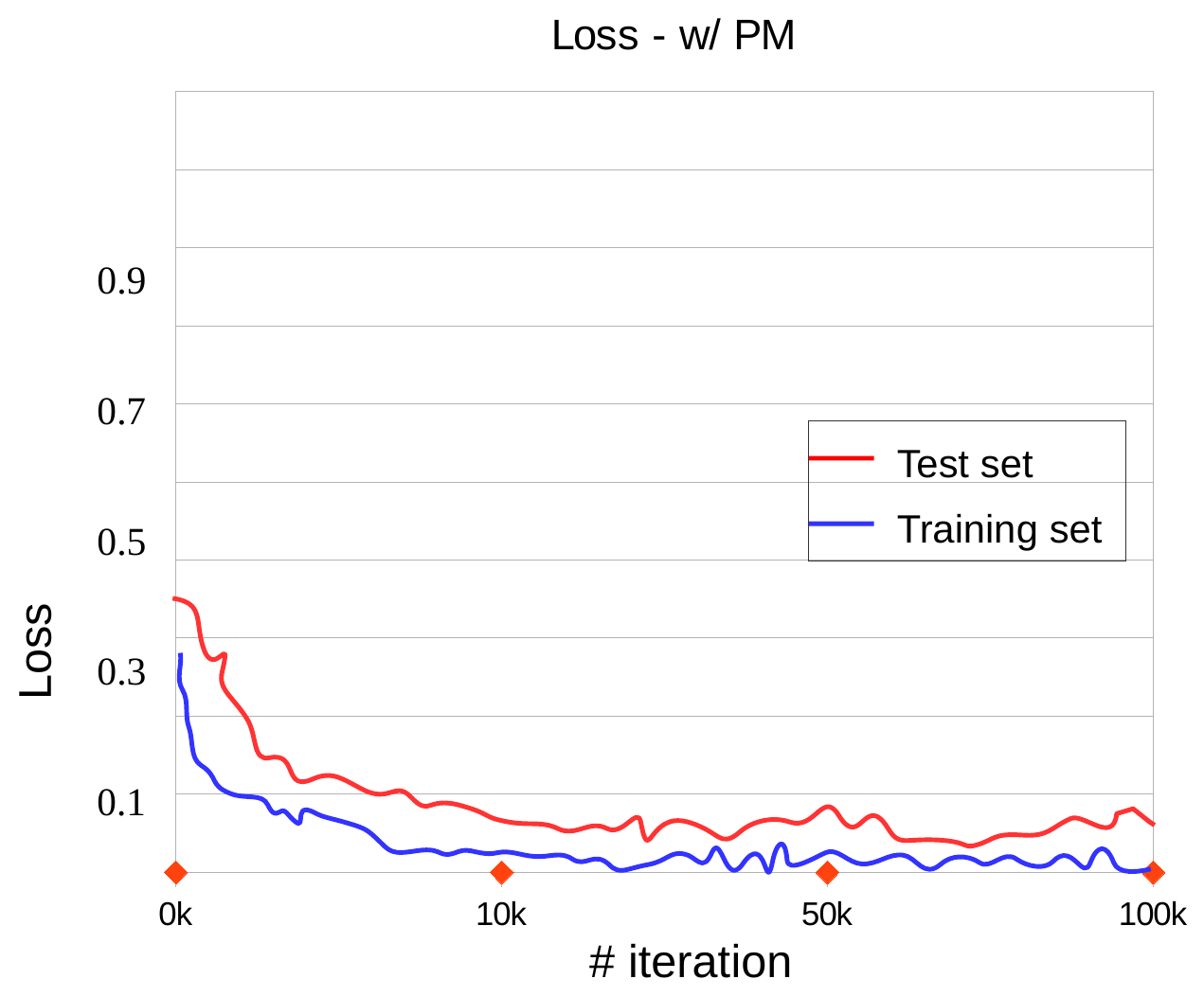}
        \includegraphics[height=3.6cm]{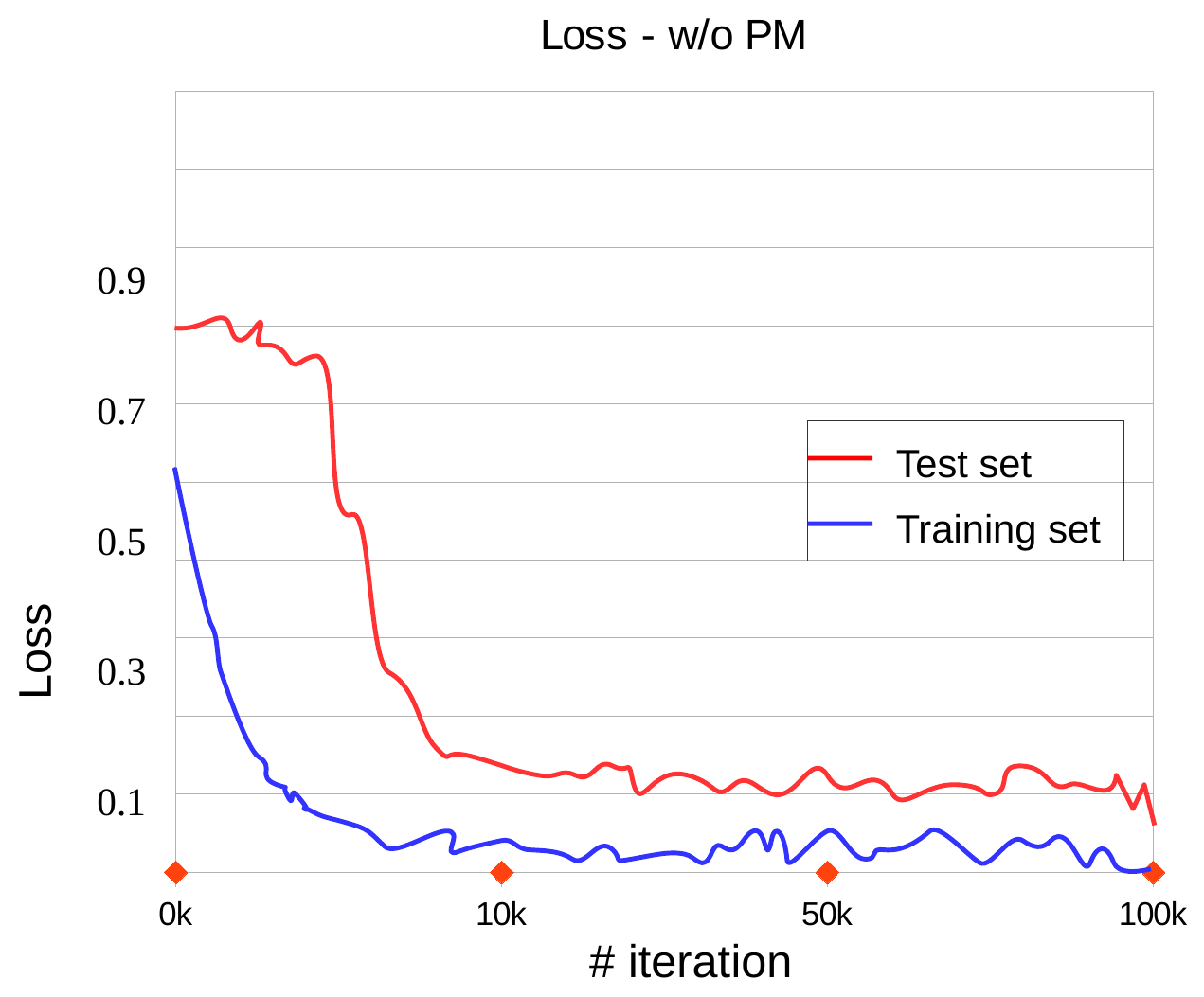}\\
  \end{tabular}\caption{Component analysis of positive sample mining. Left: CMC curves with or without positive sample mining. PM: local positive mining. NM: hard negative mining. Middle and Right: The loss of the training and test sets with/without positive sample mining, respectively. }\label{fig:ft}
\end{figure*}

We evaluate the contribution of positive sample mining by comparing the performance with and without it. Since the embedding of CRBMs which perform a series of convolution and probabilistic pooling is unsupervised encoding, it is lack of capability to maintain identities with large visual variations. Thus, positive mining in local range and the derived objective are proposed to enhance the embedding of CRBMs by reducing intra-personal variations. We performed experiments on CUHK01 data set. The left figure of Fig. \ref{fig:ft} shows the CMC curves. We can find that the collaboration of hard negative mining and positive mining achieves the best result at rank-1 value of 73.53\%. Hard negative mining alone \footnote{For each positive pair, \eg $(i,j)$, their hard negatives are retrieved by selecting the samples with the biggest similarity scores: $k=\arg\max_{i,k\in
\bar{N}} S_{i,k}$, $l=\arg\max_{j.l\in
\bar{N}} S_{j,l}$.} has a performance drop to 66.48\%. This validates the positive role of positive sample mining in learning features adaptive to local manifolds. In the case of no mining method is used, the embedding gives very low identification rate at rank-1 of 56\%. The middle and right figures of Fig. \ref{fig:ft} show the loss of the training and test sets with and without positive mining. We can see that mining on local positives enables faster convergence because of its manipulation on the local feature structure. 

\begin{figure*}[t]
\begin{tabular}{ccc}
 \hspace{-1cm}  \includegraphics[height=3cm]{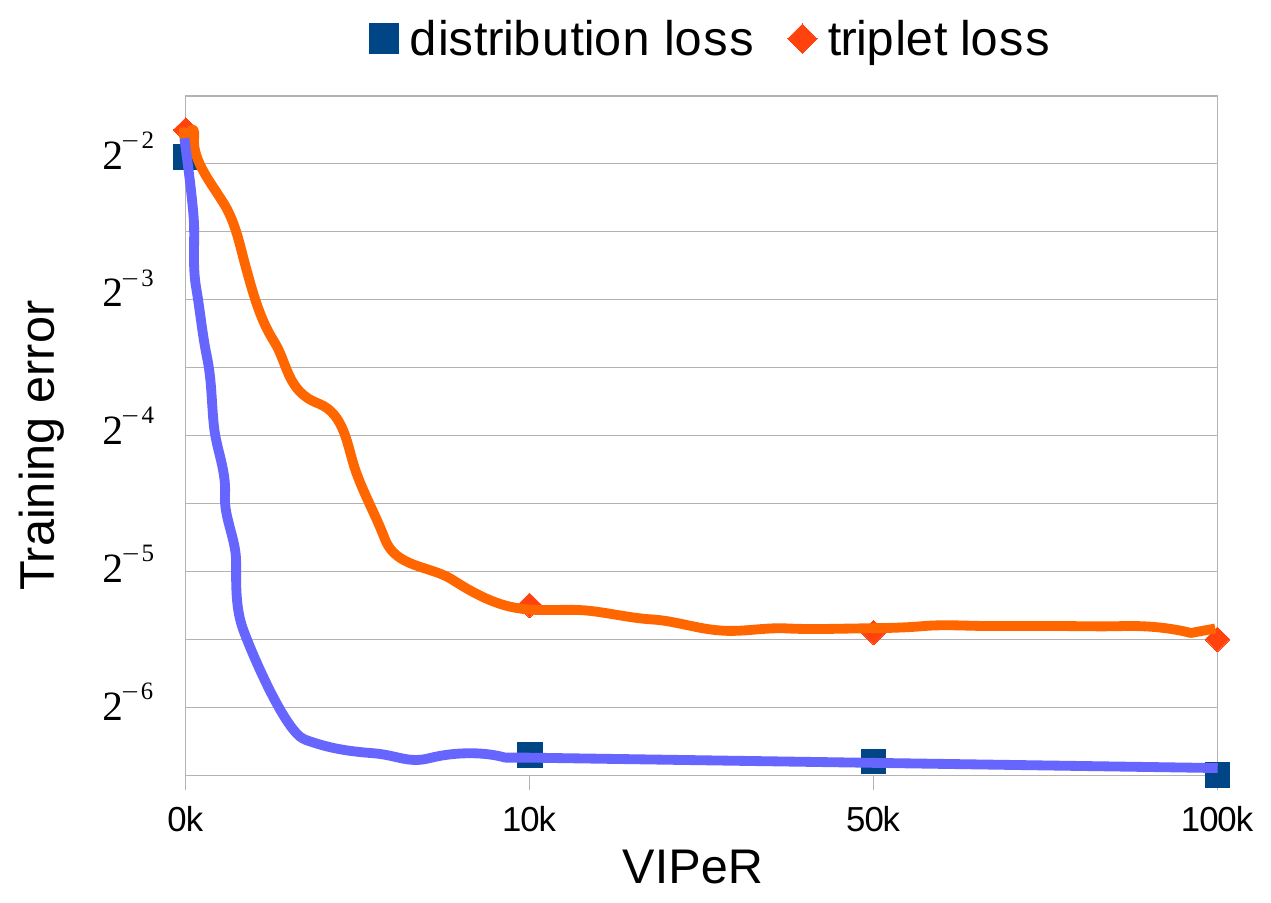}
\includegraphics[height=3cm]{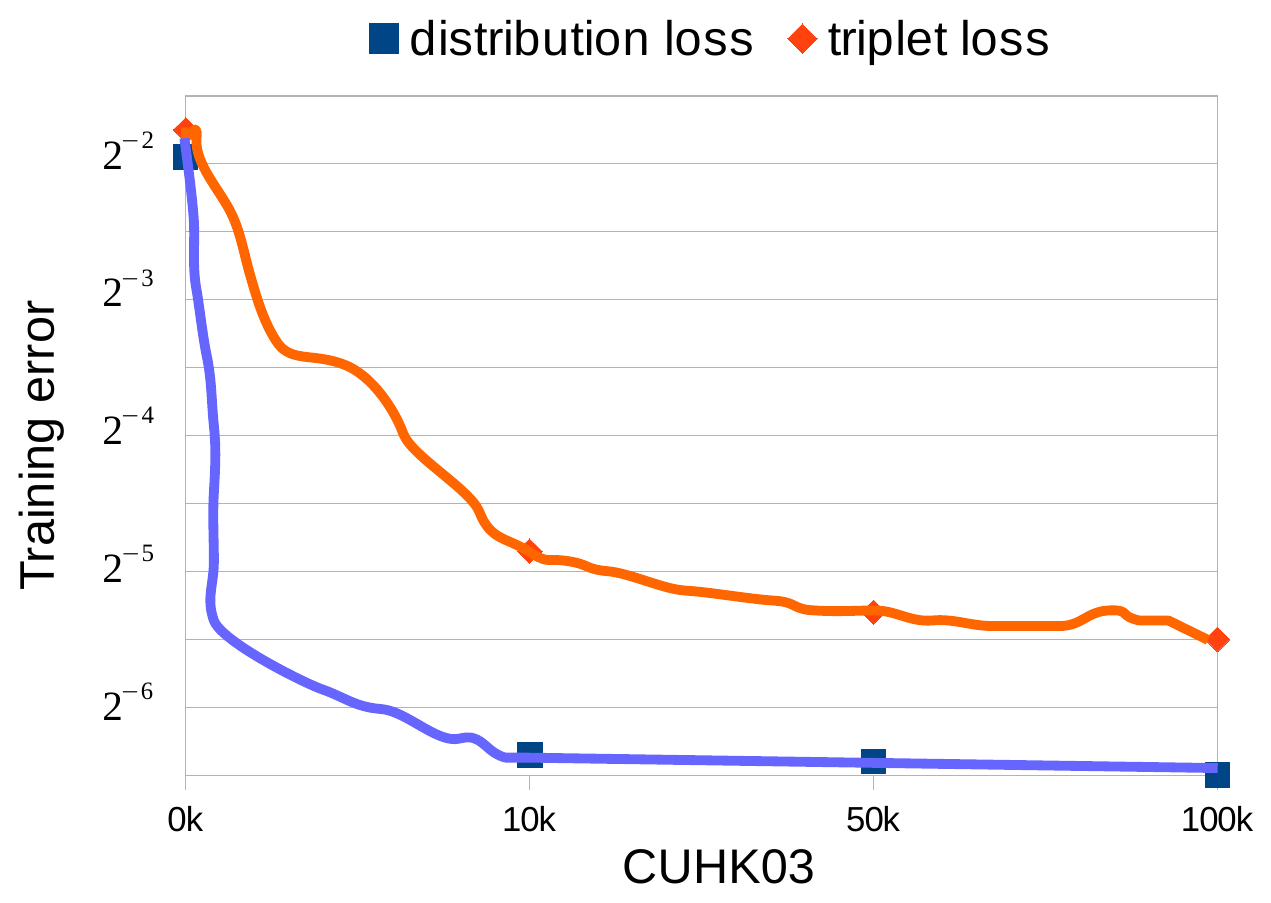}
\includegraphics[height=3cm]{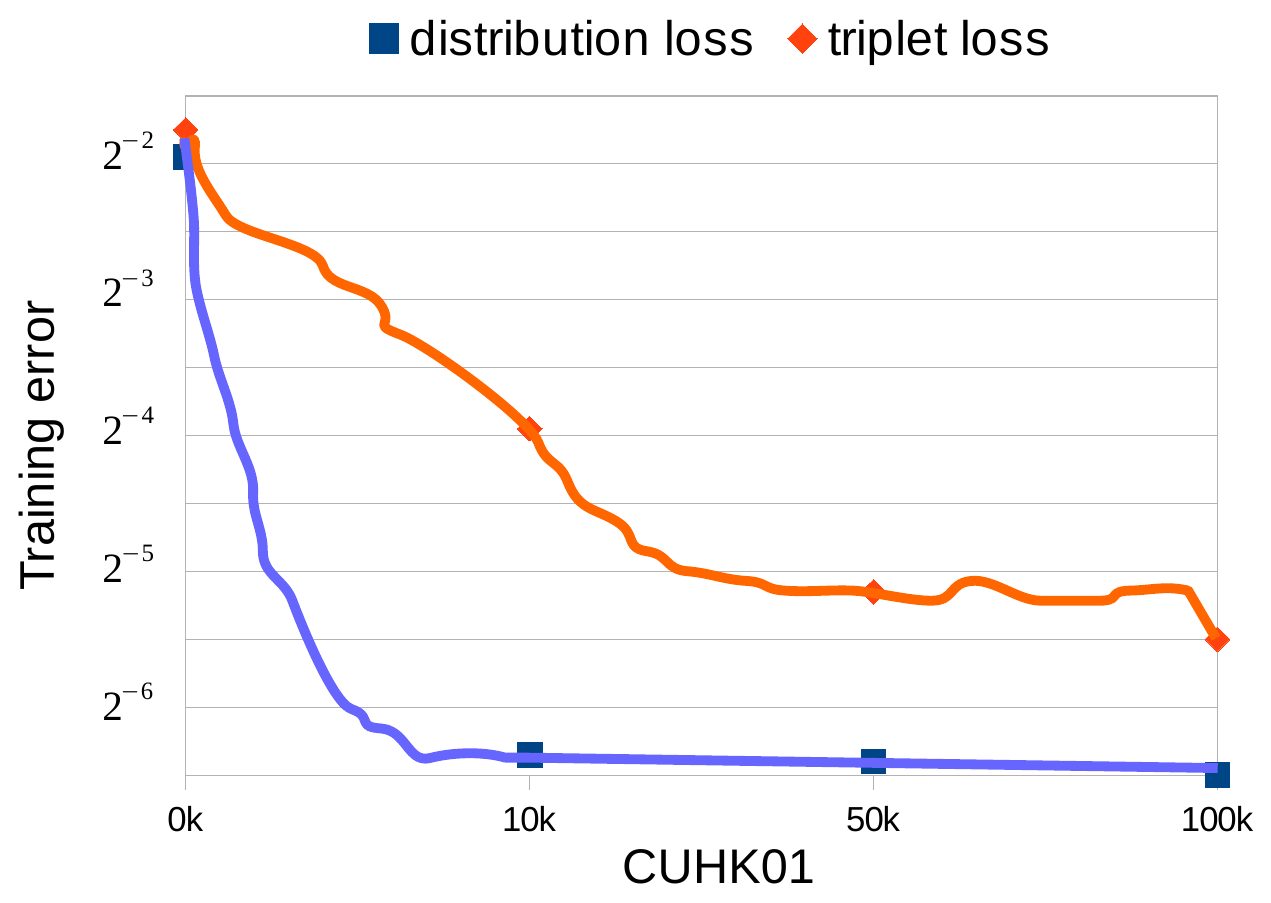}
\end{tabular}\caption{Training curves on three datasets as function of number of iterations. It indicates that local distribution loss reaches the same error to triplet loss in 5-30 times fewer iterations.}\label{fig:convergence}
\end{figure*}

\begin{figure}[t]
    \centering
        \includegraphics[height=7cm]{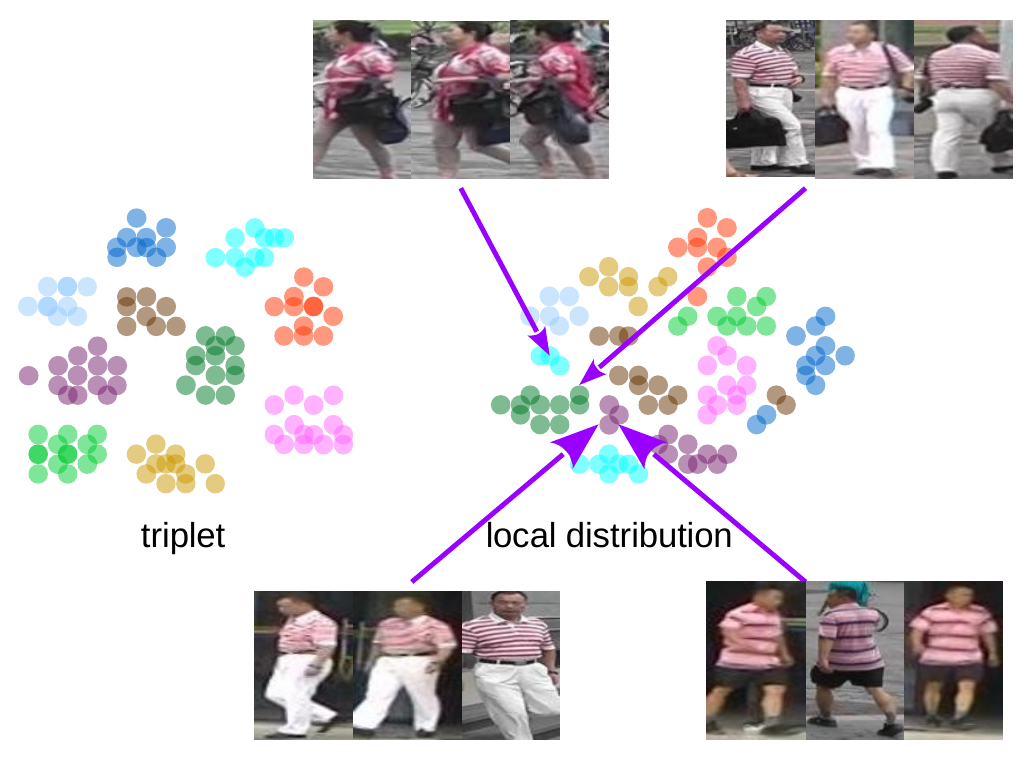}
    \caption{2D visualization of representations obtained by training triplet loss, and local distribution loss on 10 identities. Different colors correspond to different identities, and the distributions are computed from t-SNE \cite{tSNE}. Mined-out positive samples in local range from three identities are illustrated as well. See text for details.}\label{fig:visualization}
\end{figure}

\subsubsection{Convergence Rate}
We investigate the convergence rate of local distribution loss and triplet loss, and report the results in Fig. \ref{fig:convergence}. The triplet loss is implemented by using Eq.\eqref{eq:triplet_loss}, and the architecture of deep embedding is identical to our method. It can bee observed that local distribution loss can reach the asymptotic error rate of triplet loss  up to 30 times faster. Triplet loss exhibits prohibitively slow convergence in deep net training because it comes with cubic growth of the number of triplets. By contrast, distribution loss achieves efficiency by operating on meaningful positive samples in local range, leading to fewer pairwise distance comparison.

\subsubsection{Visualization on Deep Representations}
The representations are learned by feature embedding through stacked CRBMs, and optimize the objective function with similarity metric defined adaptively to local range. This dynamic process leads to more flexible representations that allow intra-personal variations, and robust to maintain identity information. To this end, 2d visualizations of representations of training with different loss functions are given in Fig. \ref{fig:visualization}. It can be seen that triplet loss tends to produce unimodal separation due to the enforcement of semantic similarity in a global Euclidean distance. By contrast, the local distribution loss can more adaptively accept intra-class variations. For example, it captures the intra-class variation between the same person holding a handbag in one camera view while without a handbag in a different view. Some mined-out positive samples in local range are also shown in Fig.\ref{fig:visualization}. These positives are moderate difficulty than very hard negative examples.

\subsection{Experimental Results}

In this section, we compare the proposed method with the following state-of-the-art approaches: JointRe-id \cite{JointRe-id}, FPNN \cite{FPNN}, LADF \cite{LADF}, SDALF \cite{Farenzena2010Person}, eSDC \cite{eSDC}, KISSME \cite{Kostinger2012Large}, kLFDA \cite{Xiong2014Person}, ELF \cite{Gray2008Viewpoint}, PCCA \cite{PCCA}, SalMatch \cite{Zhao2013SalMatch}, MLF \cite{MidLevelFilter}, DML \cite{DeepReID}, ITML \cite{Davis2007Information}, DeepRanking \cite{DeepRanking}, Multi-channel \cite{Multi-channel-part}, NLML \cite{LocalMetric}, NullRe-id \cite{NullSpace-Reid}, LMNN \cite{Hirzer2012Person}, PersonNet \cite{PersonNet}, DomainDropout \cite{DomainDropout}, LOMO+XQDA \cite{LOMOMetric}, E-Metric \cite{E-metric}, SI-CI \cite{SI-CI}, GatedCNN \cite{GatedCNN}, DeepLDA \cite{LDAFisherVector}, SSM \cite{SSM}. Note that not all of these methods report their results in all three data sets, and for fair comparison, we conduct performance with aforementioned method if available.

\begin{table}[hbt!]\small
  \centering
  \caption{Rank-1, -5, -10, -20 recognition rate of various methods on the VIPeR data set (test person =316). }  \label{tab:cmc_viper}
  {
  \begin{tabular}{c|c|c|c|c}
  \hline
\hline
    Method  & $R=1$  &  $R=5$ & $R=10$  & $R=20$ \\
  \hline\hline
   JointRe-id \cite{JointRe-id} & 34.80 & 63.32  & 74.79 & 82.45  \\
   LADF \cite{LADF}  & 29.34 & 61.04 & 75.98 & 88.10\\
   SDALF \cite{Farenzena2010Person} & 19.87 & 38.89 & 49.37 & 65.73\\
   eSDC \cite{eSDC} & 26.31 & 46.61 & 58.86 & 72.77\\
   KISSME \cite{Kostinger2012Large} & 19.60 & 48.00 & 62.20 & 77.00\\
   kLFDA \cite{Xiong2014Person} & 32.33 & 65.78 & 79.72 & 90.95\\
  ELF \cite{Gray2008Viewpoint} & 12.00 & 41.50 & 59.50 & 74.50\\
  PCCA \cite{PCCA} & 19.27 & 48.89 & 64.91 & 80.28\\
  SalMatch \cite{Zhao2013SalMatch} & 30.16 & 52.00 & 62.50 & 75.60 \\
  MLF \cite{MidLevelFilter} & 29.11 & 52.00 & 65.20 & 79.90\\
  DML \cite{DeepReID} & 34.49 & 60.13 & 74.37 & 84.18\\
  DeepRanking \cite{DeepRanking} & 38.37 & 69.22 &81.33 & 90.43\\
 Multi-channel \cite{Multi-channel-part} & 47.80 & 74.70 & 84.80 & 91.10\\
 NLML \cite{LocalMetric} & 42.30 & 70.99 & 85.23 & 94.25\\
 NullRe-id \cite{NullSpace-Reid} & 42.28 & 71.46 & 82.94 & 92.06\\
 SCSP \cite{SimilaritySpatial} & 53.54 & 82.59 & 91.49 & 96.65\\
 SI-CI \cite{SI-CI} &35.75 & 72.33 & 81.78 & 97.07\\
 E-Metric \cite{E-metric} & 40.91& 73.80 & 85.05 & 92.00\\
 GatedCNN \cite{GatedCNN}& 37.80 & 66.90 & 77.40 &-\\
 S-LSTM \cite{S-LSTM} & 42.40 &68.70  & 79.40 & -\\
\hline
   Ours &  49.04  & 77.13 & 86.26 &96.20\\
  \hline
  \end{tabular}
  }
\end{table}

\begin{table}[hbt!]\small
  \centering
  \caption{Rank-1, -5, -10, -20 recognition rate of various methods on the CUHK03 data set (test person =100).}  \label{tab:cmc_cuhk03}
  {
  \begin{tabular}{c|c|c|c|c}
  \hline
\hline
    Method  & $R=1$  &  $R=5$ & $R=10$  & $R=20$ \\
  \hline\hline
   JointRe-id \cite{JointRe-id}  & $54.74$ & 86.42 & 91.50 & 97.31 \\
   FPNN \cite{FPNN} & $20.65$ & 51.32 & 68.74 & 83.06\\
   NullRei-d \cite{NullSpace-Reid}  & $58.90$ & 85.60 & $92.45$ & 96.30 \\
   ITML \cite{Davis2007Information} & $5.53$ & 18.89 &29.96 & 44.20 \\
   LMNN \cite{Hirzer2012Person} & $7.29$ & 21.00 & 32.06& 48.94 \\
   LDM \cite{Guillaumin2009Isthatyou} & $13.51$ & 40.73 &52.13 & 70.81 \\
   KISSME \cite{Kostinger2012Large} & $14.17$ & 48.54 &52.57 & 70.03 \\
   kLFDA \cite{Xiong2014Person} & 48.20 & 59.34 & 66.38 & 76.59\\
   LOMO+XQDA \cite{LOMOMetric} & 52.20 & 82.23 & 92.14 & 96.25\\
   PersonNet \cite{PersonNet} & 64.80 & 89.40 &94.92 & 98.20 \\
   DomainDropout \cite{DomainDropout} & 72.60 & 91.00  & 93.50 & 96.70 \\
   E-Metric \cite{E-metric} & 61.32 & 89.80 & 96.50 & 98.50\\
   SI-CI \cite{SI-CI} & 52.27 & 83.45 & 85.02 & 95.68\\
   GatedCNN \cite{GatedCNN}& 61.80 & 80.90 & 88.30 & 92.20\\
\hline
  Ours &  \color{red}$\mathbf{73.02}$ & \color{red}$\mathbf{91.57}$ &   \color{red}$\mathbf{96.73}$ &  \color{red} $\mathbf{98.58}$ \\
  \hline
  \end{tabular}
  }
\end{table}

\begin{table}[hbt!]
  \centering
  \caption{Rank-1, -5, -10, -20 recognition rate of various methods on the CUHK01 data set (test person =100).}  \label{tab:cmc_cuhk01}
  {
  \begin{tabular}{c|c|c|c|c}
  \hline
\hline
    Method  & $R=1$  & $R=5 $& $R=10 $ & $R=20$ \\
  \hline\hline
   JointRe-id \cite{JointRe-id}  & $65.00$ & $88.70$ &  93.12 & 97.20 \\
   SDALF \cite{Farenzena2010Person} & 9.90 & 41.21 & 56.00 & 66.37 \\
   FPNN \cite{FPNN} & 27.87 & 58.20 & 73.46 & 86.31 \\
   LMNN \cite{Hirzer2012Person} & 21.17 & 49.67 & 62.47 & 78.62 \\
   ITML \cite{Davis2007Information} & 17.10 & 42.31 & 55.07 & 71.65 \\
   eSDC \cite{eSDC} & 22.84 & 43.89 & 57.67 & 69.84 \\
   KISSME \cite{Kostinger2012Large}  & 29.40 & 57.67 & 62.43 & 76.07 \\
   kLFDA \cite{Xiong2014Person} & 42.76 & 69.01 & 79.63 & 89.18\\
  SI-CI \cite{SI-CI} & 71.80 & 91.35 & 94.69 & 97.06\\
  SalMatch \cite{Zhao2013SalMatch} & 28.45 & 45.85 & 55.67 & 67.95\\
  E-Metric \cite{E-metric} & 69.38 & 92.16 & 96.05 & 97.00\\
\hline
Ours  & \color{red}$\mathbf{71.60}$ & \color{red}$\mathbf{93.24}$ &  \color{red}$\mathbf{96.46}$ & \color{red}$\mathbf{97.25}$ \\
  \hline
  \end{tabular}
  }
\end{table}

\begin{table}[t]\small
  \centering
  \caption{Rank-1,-5, -10, -20 recognition rate and  mAP of various methods on the Market-1501 data set (test person =751). MQ: multiple query.} \label{tab:cmc_market}
  {
  \begin{tabular}{l|c|c|c|c|c}
  \hline\hline
    Method  & $ R=1$  & $ R=5$ & $ R=10$ & $ R=20$&  mAP \\
  \hline\hline
   SDALF \cite{Farenzena2010Person} & 20.53 &- &- &- & 8.20\\
   eSDC \cite{eSDC} & 33.54 &- &- & -& 13.54\\
   KISSME \cite{Kostinger2012Large} & 39.35 &- &- &- & 19.12\\
   kLFDA \cite{Xiong2014Person} & 44.37 & 67.35 & 74.82 & 81.94 &  23.14 \\
   LOMO+XQDA \cite{LOMOMetric} & 43.79 & 65.27 & 73.22 & 80.38 & 22.22\\
   BoW \cite{Market1501} & 34.40 &- &- &- &  14.09\\
   DeepLDA \cite{LDAFisherVector} & 48.15 & 72.46 &80.22 &86.78 &  29.94\\
   Deep-Hist-Loss \cite{Deep-hist-loss} & 59.47 & 80.73 & 86.94 & 91.09 & -\\
   SCSP \cite{SimilaritySpatial} & 51.90 & -&- &- & 26.35\\
   NullRe-id \cite{NullSpace-Reid}  & 55.43 & -& -&- & 29.94\\
   NullRe-id \cite{NullSpace-Reid} + MQ & 71.56 &- &- & -& 46.03\\
   GatedCNN \cite{GatedCNN} & 65.88 &- & -& -& 39.55\\
   GatedCNN \cite{GatedCNN} + MQ & 76.04 & -& -&- & 48.45\\
   S-LSTM \cite{S-LSTM} + MQ & 61.60 & -&-&-& 35.30\\
   SSM \cite{SSM} + re-ranking & 82.81 &-&-&-& 68.80\\
   K-reciprocal encoding \cite{Re-ranking} &77.11 &-&-&-& 63.63\\
\hline
Ours & 68.32 & 87.23 & 94.59 & 96.71& 40.24 \\
Ours + MQ & \color{red}$\mathbf{84.14}$ & 93.25 & 97.33  & 98.07 & 58.80 \\
  \hline
  \end{tabular}
  }
\end{table}

\begin{figure*}[t]
   \begin{tabular}{ccc}
     \hspace{-1.2cm}    \includegraphics[height=3.8cm]{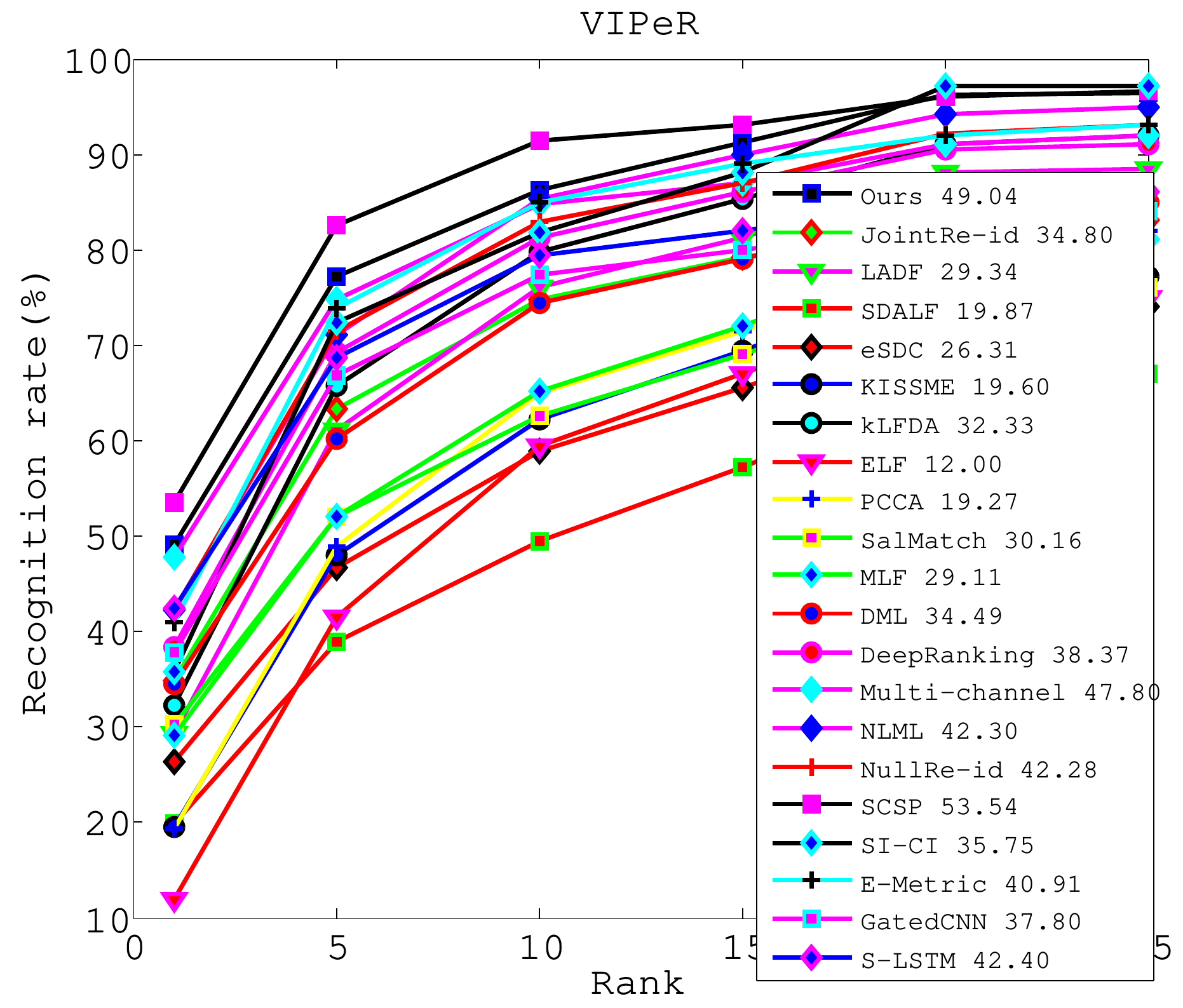}
	\includegraphics[height=3.8cm]{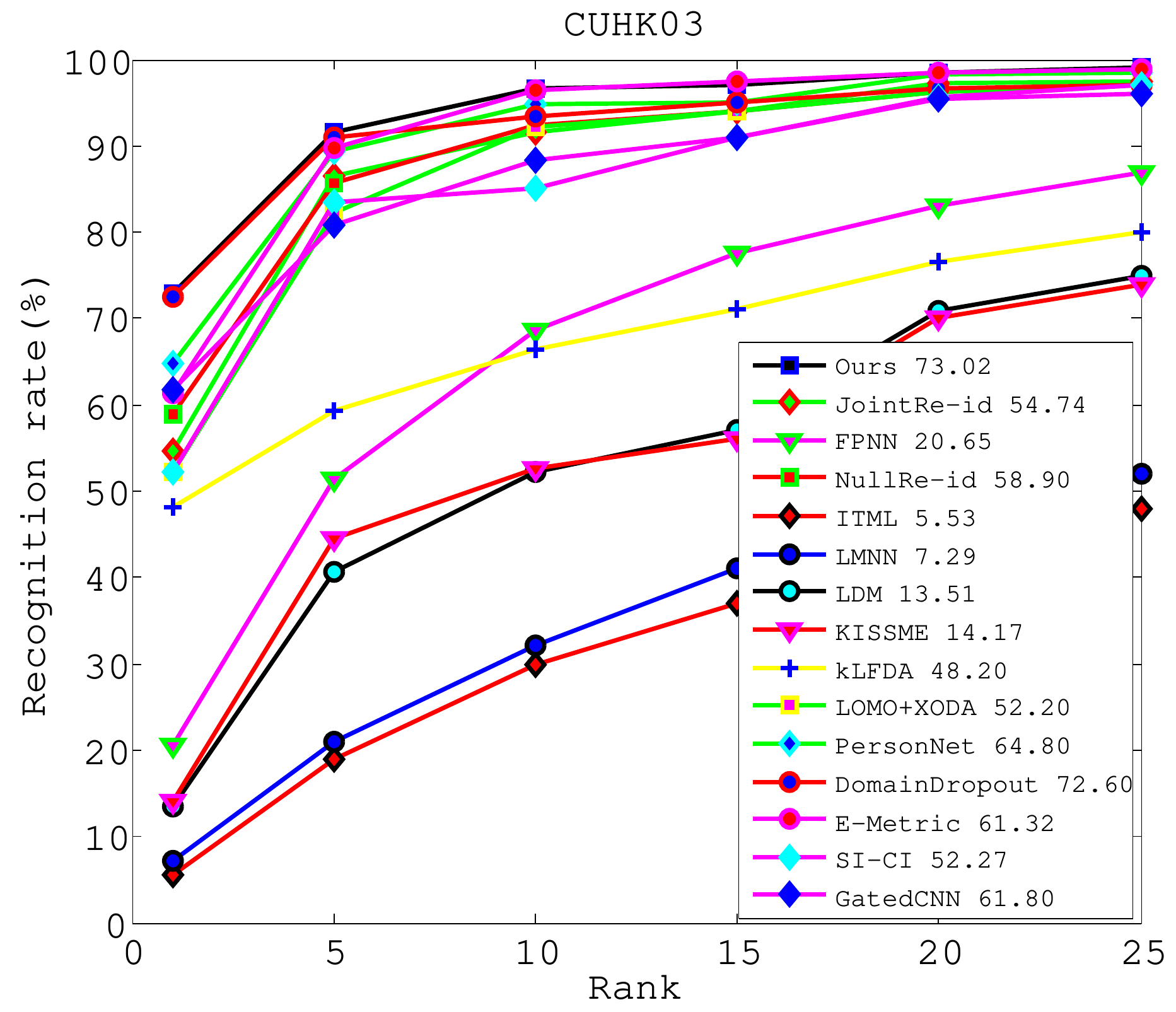}
        \includegraphics[height=3.8cm]{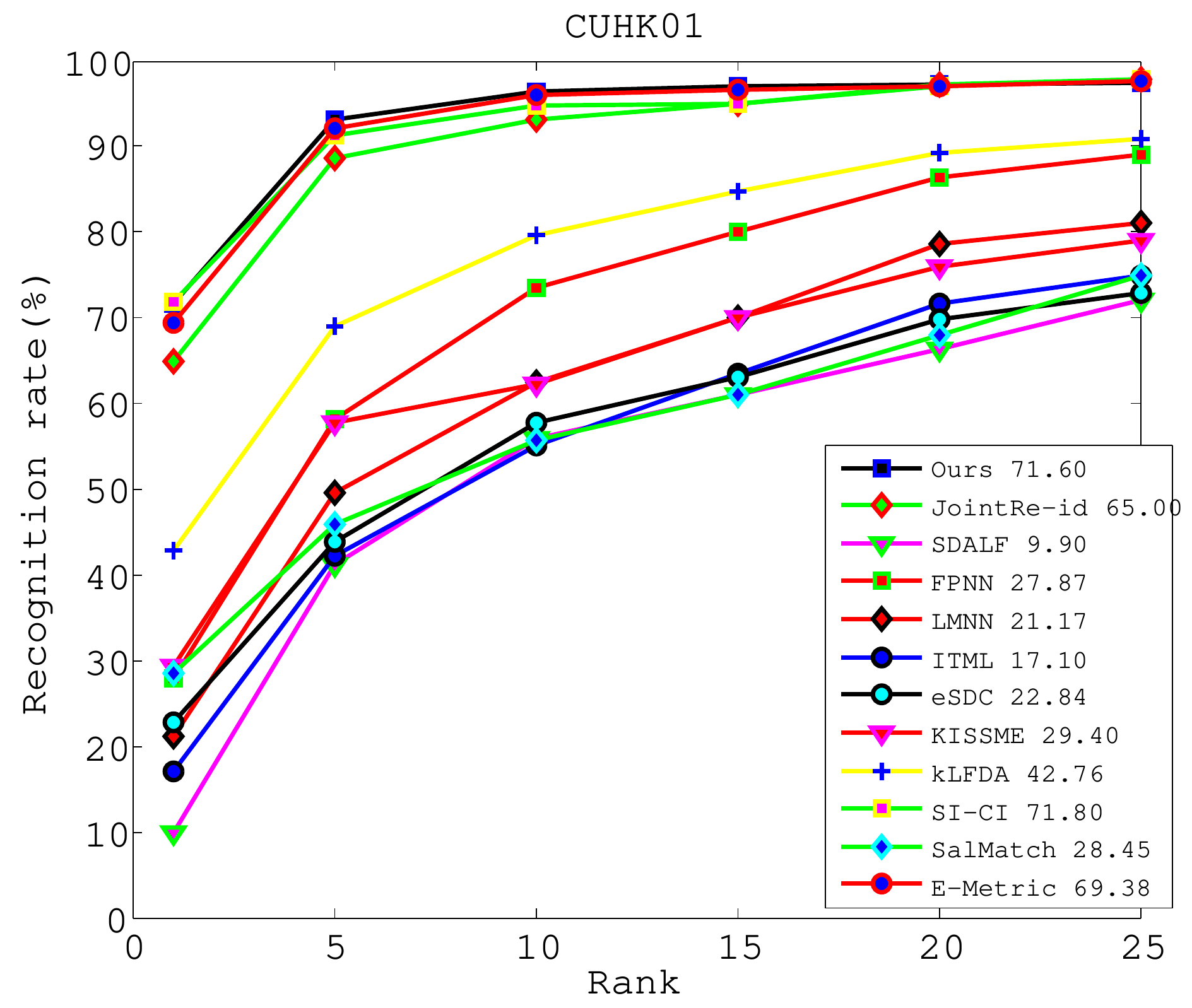}\\
\end{tabular}\caption{Performance comparison with state-of-the-art approaches using CMC curves on VIPeR, CUHK03 and CUHK01 data sets.}\label{fig:cmc}
\end{figure*}

\subsubsection{Experiments on VIPeR data set}
We compare the proposed approach to state-of-the-art methods in terms of CMC values on VIPeR dataset, and report results in Table \ref{tab:cmc_viper} and Fig.\ref{fig:cmc}. This dataset is relatively small and the number of distinct identities as well as positive pairs per identity for training are very less compared to other datasets. Therefore, we adopt a random translation for the training data augmentation. The images are randomly cropped (0-5 pixels) in horizon and vertical, and stretched to recover the size. Even though our method does not achieve the best matching rate, it outperforms recent deep embedding approaches: JointRe-id \cite{JointRe-id}, DeepRanking \cite{DeepRanking}, Multi-channel \cite{Multi-channel-part}, NLML \cite{LocalMetric}, and SI-CI \cite{SI-CI}. These methods are trained on triplet loss or soft-max loss and their similarity is defined in a global Euclidean distance. For example, our method has performance gain from 47.80\% (the best result of deep embedding in state-of-the-art) to 49.04\%. The most similar approach to us is E-Metric \cite{E-metric} (rank-1 rate is 40.91\%) which also considers moderate positive mining in local manifold structure. However, our method performs embedding with absolute feature position in deep feature space and the positive mining strategy is different from E-Metric \cite{E-metric} in the selection of hard quadruplets.

\subsubsection{Experiments on CUHK03 data set} Table \ref{tab:cmc_cuhk03} and Fig. \ref{fig:cmc} provide CMC results attained from all methods. Our method outperforms all competitors including deep embedding alternatives: PersonNet \cite{PersonNet}, DomainDropout \cite{DomainDropout}, GatedCNN \cite{GatedCNN}, E-Metric \cite{E-metric} and SI-CI \cite{SI-CI}. Compared with PersonNet \cite{PersonNet} that uses more weight layers training on pairwise sampling and joint similarity learning, our method offers advantage in jointly optimizing feature embedding, similarity metric learning, and adaptive positive sampling in local range. Compared with DomainDropout \cite{DomainDropout} which combines training samples from multiple domains to seek generic features, our approach mitigates the cross-domain problem by maintaining the distribution of samples, and results in adaptive representations.

\subsubsection{Experiments on CUHK01 data set}
The CUHK01 data set contains 971 subjects, each of which has 4 images under two camera views. Following the protocol in \cite{GenericMetric}, the data set is split into 871 subjects as training and the rest 100 as test set. Since this data set has limited number of training samples regarding each identity, we optimize the embedding on CUHK03, and then fine-tuning on the objective function on CUHK01. The CMC rank rates and curves are shown in Table \ref{tab:cmc_cuhk01} and Fig.\ref{fig:cmc}, our method consistently achieves performance gain by improving the state-of-the-art result from 69.38\% (attained by E-Metric \cite{E-metric}) to 71.60\% at rank-1 matching rate. This notable improvement can be credited to the expressive representations derived from the deep embeddings of our method. The proposed joint optimization on embedding, similarity metric learning, and more suitable positive mining improves the quality of representation learning in person re-id.

\subsubsection{Experiments on Market-1501 data set}

This dataset is the largest and the most realistic dataset with natural detector errors abundant in the provided data. Since each subject is captured by 6 cameras, the intra-class variations are evidently dominating in samples therein. In this data set, our method outperforms most of competitors in terms of CMC rank-1 rate and the mean average precision (mAP) value. The results are shown in Table \ref{tab:cmc_market}. In the setting of single query, a number of alternatives are based on classical metric learning, \eg KISSME \cite{Kostinger2012Large}, kLFDA \cite{Xiong2014Person}, LOMO+XQDA \cite{LOMOMetric}, and NullRe-id \cite{NullSpace-Reid}, whereas they are unable to jointly optimize feature embedding and similarity learning. Our approach achieves 68.32\% at rank-1 rate and 40.24\% as mAP, showing a notable margin against these methods in single query case. It can be observed that two state-of-the-arts \ie SSM \cite{SSM} + re-ranking and K-reciprocal encoding \cite{Re-ranking} outperform our method in single query setting. The main reason is SSM \cite{SSM} and k-reciprocal encoding \cite{Re-ranking} rely on a re-ranking process, which is not a self-contained principle. While deep embedding methods of DeepLDA \cite{LDAFisherVector}, Deep-Hist-Loss \cite{Deep-hist-loss}, SCSP \cite{SimilaritySpatial} and GatedCNN \cite{GatedCNN} can improve their recognition accuracy to some extend, they commonly perform embeddings with a global Euclidean objective. In contrast, our approach achieves a better result on account of jointly optimizing feature embeddings, adaptive similarity learning and local-ranged positive mining. In the setting of multiple query (MQ), the proposed method outperforms the most state-of-the-arts \ie NullRe-id \cite{NullSpace-Reid} + MQ, GatedCNN \cite{GatedCNN} + MQ, S-LSTM \cite{S-LSTM} + MQ, by a notable margin. The witnessed performance gain can be ascribed to the reduced intra-class variations yielded by our method with respect to the multiple query structure.

\subsection{More Evaluations and Analysis}

In this section, we carry out a fair self-evaluation of the proposed deep adaptive feature embedding approach to person re-identification. Fair self-evaluation is treated as evaluation of both the final output and each component of the algorithm to assess the actual contributions of therein. According to \cite{Xiong2014Person,DeepRanking}, a fair self-evaluation is supposed to verify that the effectiveness of the proposed method stems primarily from the components which are claimed to be effective, rather than comparing only the final output or a specific component in different settings. Our aim to conducting a self-evaluation is to assess each component of our approach to prove the necessarily positive roles that all components have played in the whole framework.

\subsubsection{Evaluation on Hierarchical Representations: CNN features v.s. CRBMs}

In this experiment, we test the capacity of our method in learning hierarchical representations for pedestrian images. In the embedding stage, we are using a stacked, three-layer CRBMs, and we retain features from each layer to perform pedestrian matching. As alternative, CNN is widely adopted to be the embedding solution \cite{E-metric,SI-CI,PersonNet,GatedCNN}. However, CNN typically uses shared parameters and filters across the input and feed-forward dimensions. This is unable to learn specific features from the different human body parts of pedestrian images; meanwhile, the morphological information is preserved from each part of human body. Results are shown in Table \ref{tab:layers}. First, it can be seen that CNN embedding cannot produce representations as robust as CRBMs can. This demonstrates the rational of using CRBMs as the feature embedding. Second, in terms of individual layers, the third layer outperforms the second and the first layer, and the combination of the first and second layers notably improves the matching accuracy relative to the first layer alone but marginal to the second layer alone. Third, the combination of three layers can improve the performance to a very limited extent relative to the third layer alone. Thus, to keep computation efficiency, we use the features extracted from the third layer only.

To illustrate the effectiveness of the proposed quadruplet loss function, we perform experiments by applying a variety of loss functions into existing feature learning networks. Baselines using AlexNet \cite{AlexNet}, VGGNet \cite{VGG} and ResNet \cite{ResNet} are fine-tuned with the default parameter setting except that the output dimension of the last FC layer is set to the number of training identities. The loss functions include contrastive loss \cite{GatedCNN,S-LSTM}, triplet loss \cite{DeepRanking}, and the proposed quadruplet loss with local positive mining. The AlexNet and VGGNet are trained for 60 epochs with a learning rate of 0.001 and then for another 20 epochs with a learning rate of 0.0001. As suggested by \cite{CNN-SL-ML}, the parameters of the convolutional layers and  first two FC layers are initialized by the parameters pre-trained on ImageNet and the parameters of the last fully-connected layer are randomly initialized with a Gaussian distribution ($G(\mu,\sigma)$,$\mu=0,\sigma=0.01$).The ResNet is trained for 60 epochs with learning rate of 0.001 initially and reduced by 10 at 25 and 50 epochs. During testing, the FC6 descriptor of AlexNet/VGGNet and the Pool5 descriptor of ResNet are used for feature representation. Experimental results are shown in Table \ref{tab:evaluate_obj}. In general, feature embedding with triplet loss outperforms contrastive loss by considering the error of the relative distance between positive and negative pairs while the contrastive loss is based on binary classification mode, which emphasizes on absolute distances of positive/negative pairs and thus leads to poor generalization ability on testing data. However, a model that is trained by a triplet loss would still cause a relatively large intra-class variation, as observed in \cite{Multi-channel-part}. The comparison of the proposed loss to the triplet loss suggests that mining hard negatives together with local positives can effectively reduce intra-class variations and thus improve the recognition rate by a notable margin.

\begin{table}[t]
  \centering
  \caption{Rank-1 accuracy on the CUHK01 data set (test person = 486 or 100) using features from CNN embedding and different individual/combination layers of CRBMs.}  \label{tab:layers}
  {
  \begin{tabular}{l|c|c}
  \hline
\hline
    Method  & test = 486  & test = 100 \\
  \hline\hline
  untied CNN embedding \cite{E-metric} & 51.79\% & 69.38\%\\
  CDBN (first layer)  & 58.32\% & 61.04\%\\
  CDBN (second layer) & 63.67\% & 66.25\% \\
  CDBN (third layer) & 65.85\% & 71.60\%\\
  CDBN (first + second layers) & 66.04\% & 71.68\%\\
  CDBN (first + second + third layers) & 66.09\% & 71.70\%\\
  \hline
  \end{tabular}
  }
\end{table}

\begin{table}[t]
  \centering
  \caption{Rank-1 accuracy on the CUHK01 data set (test person = 486 or 100) using features extracted from CNNs with varied embedding functions.}\label{tab:evaluate_obj}
  {
  \begin{tabular}{l|c|c}
  \hline
\hline
    Method  & test = 486  & test = 100 \\
  \hline\hline
  AlexNet \cite{AlexNet} + contrastive loss & 41.24\% & 62.30\%\\
  AlexNet \cite{AlexNet} + triplet loss  & 56.74\% & 70.02\%\\
  AlexNet \cite{AlexNet} + proposed loss & 62.69\% & 72.14\% \\
  VGGNet \cite{VGG} + contrastive loss & 42.56\% & 64.78\%\\
  VGGNet \cite{VGG} + triplet loss  & 58.78\% & 72.45\%\\
  VGGNet \cite{VGG} + proposed loss & 65.08\% & 75.93\% \\
  ResNet \cite{ResNet} + contrastive loss & 47.28\% & 65.04\%\\
  ResNet \cite{ResNet} + triplet loss  & 60.13\% & 76.05\%\\
  ResNet \cite{ResNet} + proposed loss & 67.29\% & 80.47\% \\
  \hline
  \end{tabular}
  }
\end{table}

\subsubsection{Analysis of Variance Reduced SGD}

\begin{figure*}[t]
   \begin{tabular}{ccc}
  \hspace{-1cm}     \includegraphics[height=3.5cm]{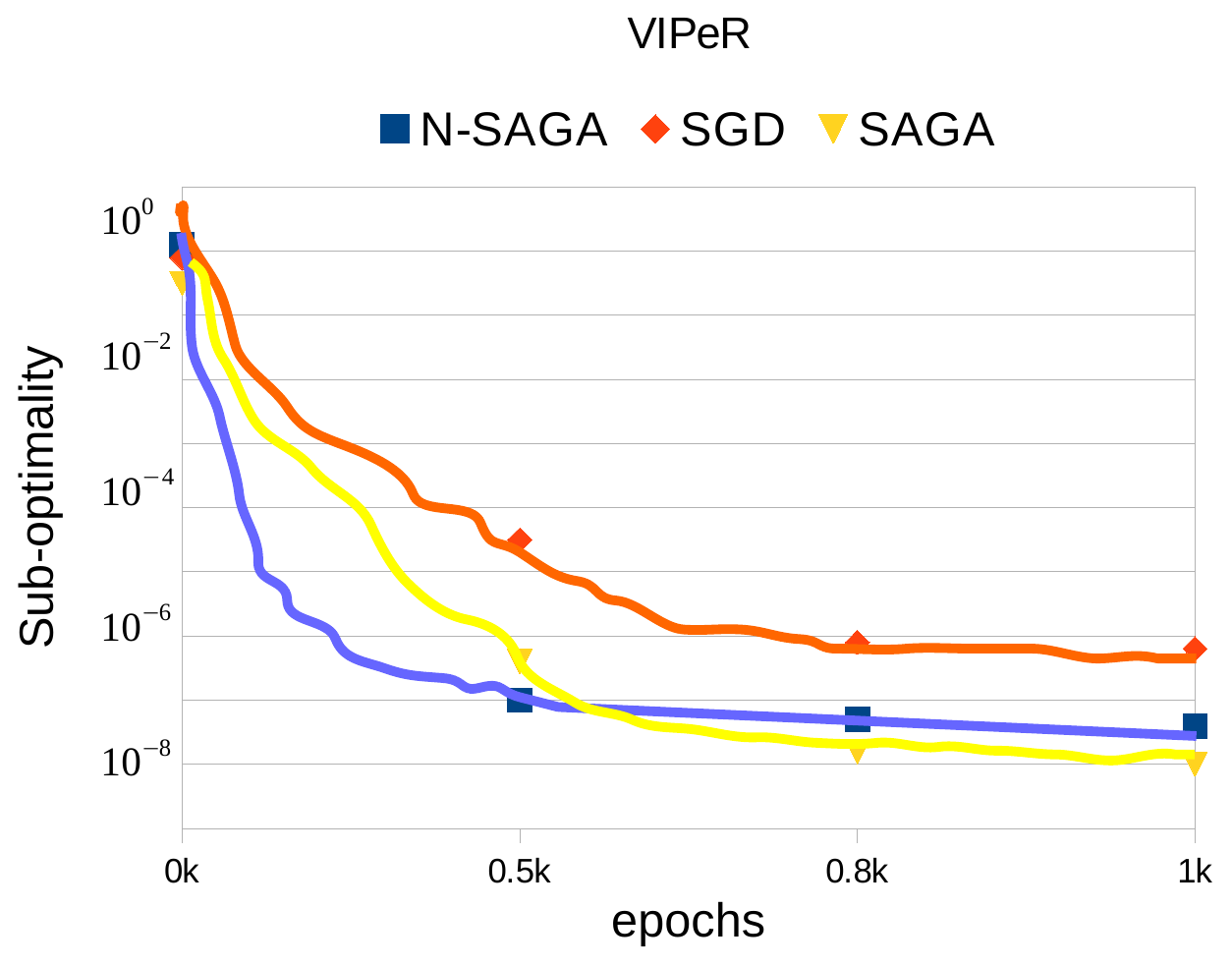}
        \includegraphics[height=3.5cm]{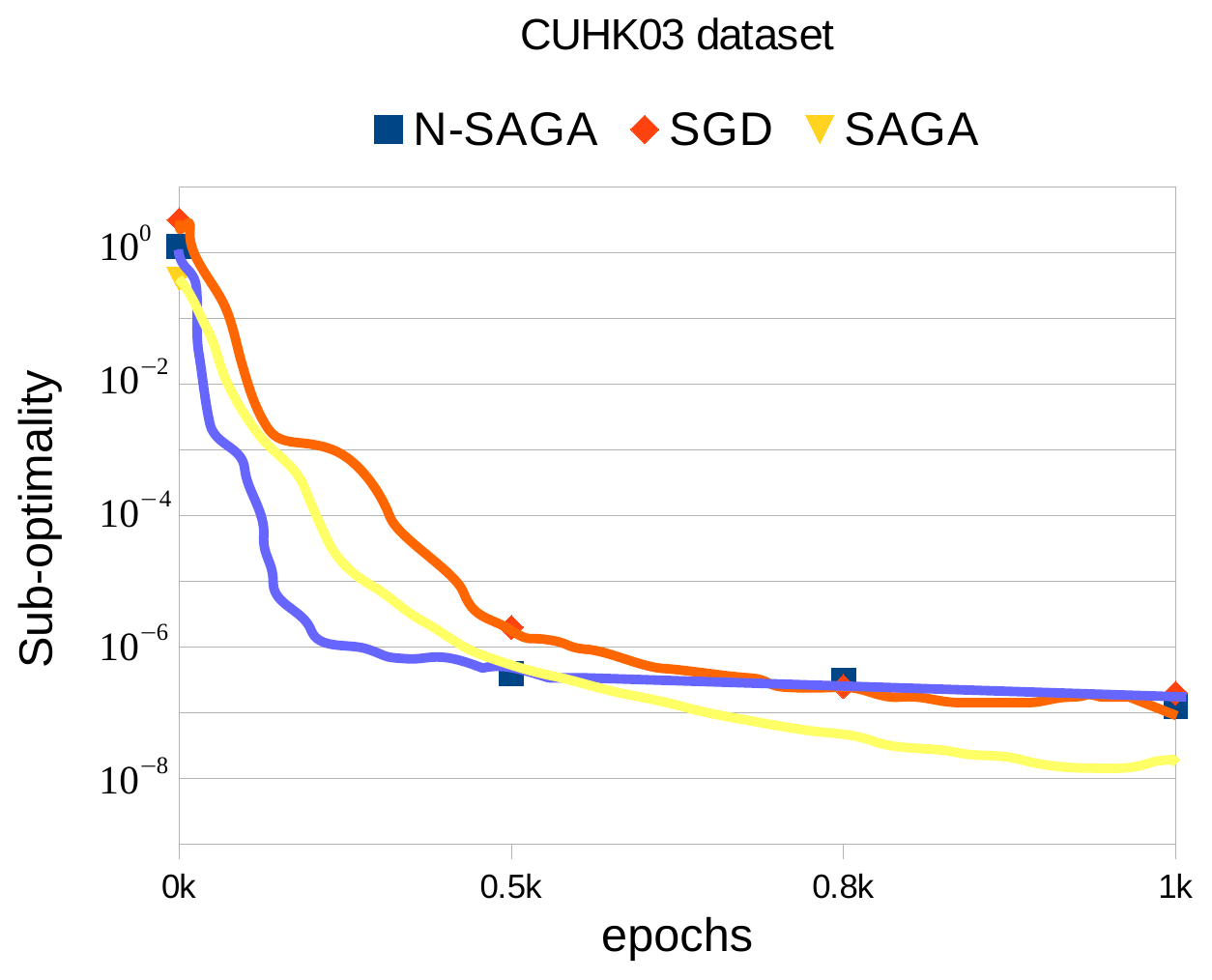}
        \includegraphics[height=3.5cm]{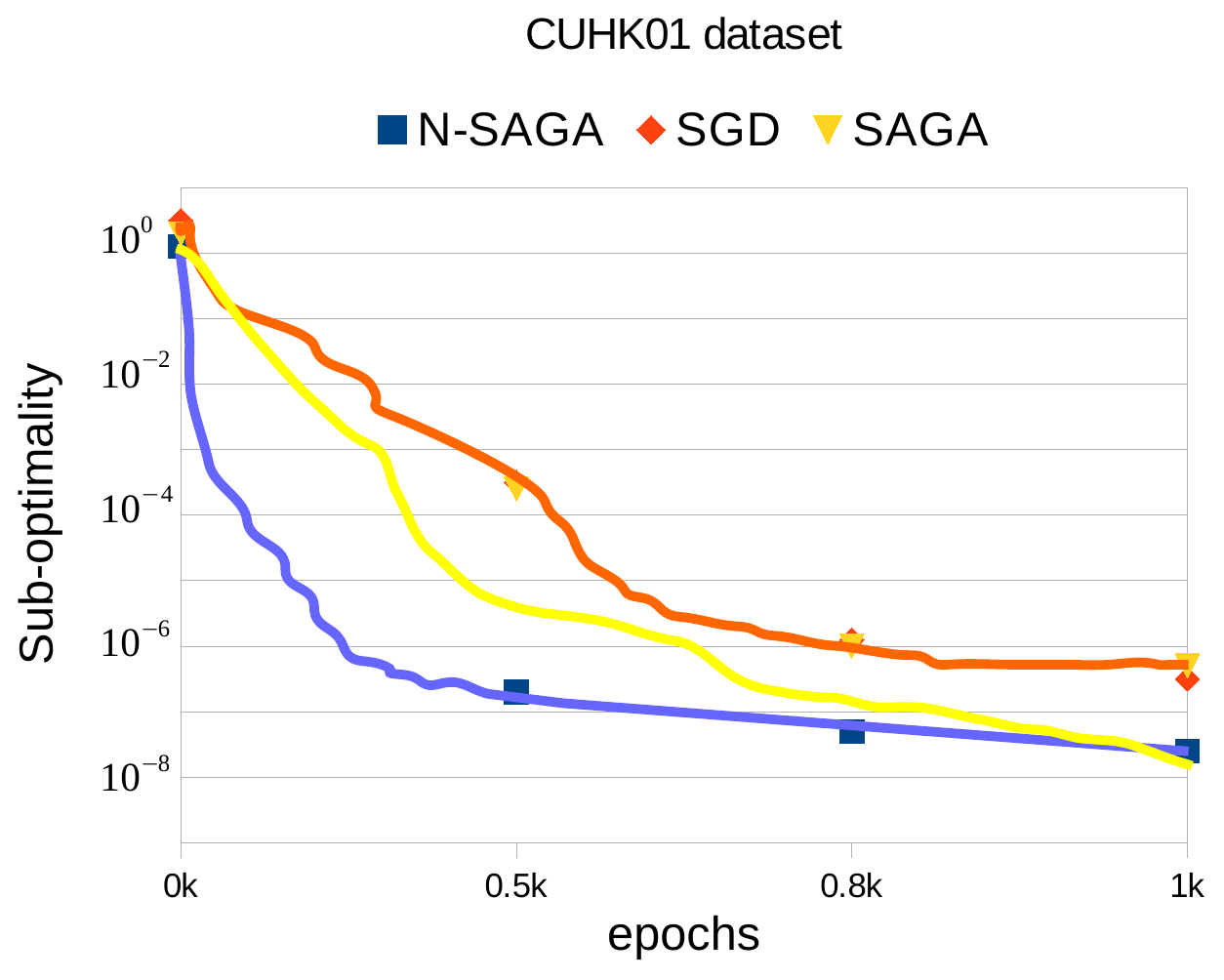}\\
 \end{tabular}\caption{Comparison of standard SGD, SAGA, and N-SAGA with decreasing and constant step size on three datasets.}\label{fig:sub_optimality}
\end{figure*}

More recent studies on variance reduction techniques such as SAGA \cite{SAGA}, N-SAGA  \cite{VRSGD} and SVRG \cite{AccelerateSGD} have been proposed to overcome this weakness, achieving linear convergence with geometric rates. In particular, they introduce variance corrections to ensure the convergence for constant learning rates. To realize the training efficiency, we employ the variance reduced SGD with sharing on stochastic gradients computed from neighborhood structure \cite{VRSGD}, and this technique is successfully applied into deep person re-id system. In this experiment, we evaluate the evolution of the sub-optimality of the objective as a function in terms of the number of update steps performed \ie data point evaluation. The step size of $\gamma=\frac{q}{\mu \mathtt{N} }$ is used. Fig.\ref{fig:sub_optimality} shows the sub-optimality as a function of the number of data point evaluations (\ie number of stochastic updates) for the value of $\mu=10^{-1}$. We can see that N-SAGA shows the minimum number of updates in stochastic computations. The constant step-size variant of SGD is faster in the early stages until it converges to a neighborhood of the optimum.

\section{Conclusion and Future Work}
In this paper, a principled deep feature embedding approach for person re-identification is presented to learn adaptive deep transformations to a feature space such that the local manifold structure of data is considered. The proposed approach is the first attempt to jointly optimize local similarity metric learning, meaningful positive mining in local manifolds and robust feature embedding. As a result, trainable parameters for feature embedding are optimized on local distribution objective which is designed to seek local manifold structure between positive samples in order to address the large intra-personal variations. To further improve training efficiency, we employ variance reduced SGD to share and reuse computed gradients across data samples in their neighborhood structure. In our future work, we would explore effective feature embeddings with discriminant analysis \cite{GTDA-TPAMI07,Geometric-TPAMI09,Large-margin-icml14} to preserve the discriminative information regarding identities while approaching a solution to stable optimization and  convergence.

\input{appendix.tex}

\bibliographystyle{elsarticle-num}
\bibliography{elsarticle-template}

\end{document}

%% file: appendix.tex
\section{Appendix: Generating Hierarchical Representations}\label{sec:appendix}

This appendix provides the procedure of generating hierarchical representations for each pedestrian sample by using CRBMs. Suppose all parameters in stacked CRBMs are learned, we can generate the representation of an image by sampling from the joint distribution over all hidden layers conditioned on the input. We use block Gibbs sampling to sample from the units of each layer in parallel.  We summarize the procedure of sampling in Algorithm \ref{alg:sampling} where we describe a case with one CRBM. The model takes image as input of visible units with whitened pixel intensity values. In person re-id, some hand-crafted features such as Local Binary Pattern (LBP) \cite{LocallyAlligned,Gheissari2006Person,Xiong2014Person} and Gabor \cite{Gray2008Viewpoint,LocallyAlligned} are employed to describe a person's appearance from different perspectives. Thus, we additionally learn deep representations from LBP and Gabor which can capture their high-order statistics. Since the generated features are high-dimensional, we use PCA to reduce the dimensionality to 500 for each type of representation.

\begin{algorithm}[t]
\KwIn{\small A visible layer (input image) $V$, a detection layer $H$, a pooling layer $P$, higher detection layer $\widehat{H}$ above $P$. A set of shared weights $\Omega=\{\Omega^{1,1},\ldots,\Omega^{K,\widehat{K}}\}$ where $\Omega^{k,l}$ is connecting pooling units $P^k$ to detection unit $\widehat{H}^l$.}
\KwOut{Hierarchical representations of an image.}
Sample from $H$ via $I(h_{ij}^k)\overset{\Delta}{=} b_k + (\tilde{W}^k \ast v)_{ij}$\;

Sample from $P$ via $I(p_{\alpha}^k) \overset{\Delta}{=}  \sum_{l} (\Omega^{k,l} \ast \widehat{h}^l)_{\alpha}$\;

\For{Sample each block independently via conditional probability}{
\small $P(h_{ij}^k=1|\mathbf{v},\mathbf{\widehat{h}})=\frac{\exp(I(h_{ij}^k) + I(p_{\alpha}^k) )}{1+\sum_{(i',j')\in B_{\alpha}} \exp(I(h_{i'j'}^k) I(p_{\alpha}^k))}$\;

$P(p_{\alpha}^k=0| \mathbf{v},\mathbf{\widehat{h}})=\frac{1}{1+\sum_{(i',j')\in B_{\alpha}} \exp(I(h_{i'j'}^k) + I(p_{\alpha}^k) )}$\;

Sample from $V$ given $H$ via Eq \eqref{eq:con_p}\;
}
\Return{Sampling results from $P$}\;
\caption{ Block Gibbs sampling on all hidden layers.}
\label{alg:sampling}
\end{algorithm}